\RequirePackage{fix-cm}
\documentclass[twocolumn]{svjour3}          
\smartqed  
\usepackage{graphicx}
\usepackage[shortlabels]{enumitem}
\usepackage{amsmath,amssymb,amsfonts}

\usepackage{graphicx}
\usepackage{textcomp}
\usepackage{xcolor}
\usepackage{textgreek}
\usepackage{pgfplots}
\pgfdeclarelayer{background}
\pgfdeclarelayer{foreground}
\pgfsetlayers{background,main,foreground}
\pgfplotsset{compat=newest}
\usepackage{tikz}
\usepackage{chemplants}
\usepgfplotslibrary{statistics}
\usetikzlibrary{shapes.geometric, arrows}

\usepackage{url}
\usepackage{breakurl}
\usepackage[breaklinks]{hyperref}

\usepackage{amsmath}
\usepackage{xcolor}
\usepackage{booktabs}
\definecolor{darkred}{rgb}{0.55, 0.0, 0.0}
\definecolor{darkgoldenrod}{rgb}{0.72, 0.53, 0.04}
\definecolor{darkkhaki}{rgb}{0.74, 0.72, 0.42}
\definecolor{navyblue}{rgb}{0.0, 0.0, 0.5}
\definecolor{moonstoneblue}{rgb}{0.45, 0.66, 0.76}
\definecolor{ashgrey}{rgb}{0.7, 0.75, 0.71}
\definecolor{blond}{rgb}{0.98, 0.94, 0.75}
\definecolor{burntumber}{rgb}{0.54, 0.2, 0.14}
\definecolor{darkcyan}{rgb}{0.0, 0.55, 0.55}
\usetikzlibrary{shapes,shadows,trees,patterns}

\usepackage{pgfplotstable}
\usetikzlibrary{pgfplots.groupplots}

\usepackage{siunitx}

\usepackage{multirow}    

\usepackage{xcolor,soul,framed} 
\usepackage{makecell}
\colorlet{shadecolor}{yellow}
\usepackage{fourier}
\usepackage[justification = centering]{caption}
\usepackage{algorithm}
\usepackage{bm}
\usepackage{romannum}
\usepackage{algpseudocode}
\pgfplotsset{select coords between index/.style 2 args={
    x filter/.code={
        \ifnum\coordindex<#1\fi
        \ifnum\coordindex>#2\fi
    }
}}

\usepackage{tikz}
\usetikzlibrary{patterns}
\usepackage{bbding}

\begin{document}
\pagenumbering{arabic}
\title{Surrogate Assisted Evolutionary Multi-objective Optimisation applied to a Pressure Swing Adsorption system 
}


\author{Liezl Stander    \and
        Matthew Woolway  \and
        Terence L. Van Zyl
}


\institute{\Envelope Liezl Stander \at
              University of Johannesburg \\
              Institute for intelligent systems \\
            \email{Liezl.Stander@sasol.com}           
            \and
            Matthew Woolway \at
            University of Johannesburg \\
            Faculty of Engineering and the Built Environment \\
            \email{mjwoolway@uj.ac.za}
            \and
            Terence L. Van Zyl \at
            University of Johannesburg \\
            Institute for intelligent systems \\
            \email{tvanzyl@uj.ac.za}
}

\date{Received: date / Accepted: date}

\maketitle

\begin{abstract}

Chemical plant design and optimisation have proven challenging due to the complexity of these real-world systems. The resulting complexity translates into high computational costs for these systems' mathematical formulations and simulation models. Research has illustrated the benefits of using machine learning surrogate models as substitutes for computationally expensive models during optimisation. This paper extends recent research into optimising chemical plant design and operation. The study further explores Surrogate Assisted Genetic Algorithms (SA-GA) in more complex variants of the original plant design and optimisation problems, such as the inclusion of parallel and feedback components. The novel extension to the original algorithm proposed in this study, Surrogate Assisted NSGA-\Romannum{2} (SA-NSGA), was tested on a popular literature case, the Pressure Swing Adsorption (PSA) system. We further provide extensive experimentation, comparing various meta-heuristic optimisation techniques and numerous machine learning models as surrogates. The results for both sets of systems illustrate the benefits of using Genetic Algorithms as an optimisation framework for complex chemical plant system design and optimisation for both single and multi-objective scenarios. We confirm that Random Forest surrogate assisted Evolutionary Algorithms can be scaled to increasingly complex chemical systems with parallel and feedback components. We further find that combining a Genetic Algorithm framework with Machine Learning Surrogate models as a substitute for long-running simulation models yields significant computational efficiency improvements, 1.7 - 1.84 times speedup for the increased complexity examples and a 2.7 times speedup for the Pressure Swing Adsorption system. It is worth noting that the SA-GA and SA-NSGA algorithms implemented in this study yielded results confirming both their equivalency, flexibility and robustness toward more complex single and multi-objective systems.

\end{abstract}

\section{Introduction}\label{intro}
In the Industrial Chemicals Industry, the optimisation of chemical processes plays a crucial role in decision-making related to process improvement and performance enhancement. Both of these activities need to be considered by engineers while attempting to reduce costs~\cite{morales2011framework}. Additionally, they must contend with a significant amount of complexity, usually involving multiple (often) conflicting objectives and numerous control variables~\cite{bernardo2001quality}. Optimisation of systems like these requires a model representing the system's behaviour~\cite{rangaiah2009multi}. A popular system representation for these models in chemical plant design is computer simulations~\cite{meyer2011innovative}. The simulations play a central role in enabling the optimisation of these systems as they are often used as the primary evaluation platform ~\cite{emmerich2001design,whitley1994genetic}. Unfortunately, optimisation across these complex system simulations often results in high turnaround times and computational costs.

When investigating the available case studies focused on chemical plant optimisation~\cite{henao2011surrogate,carpio2018enhanced,shi2016evolutionary}, there exists a wide variety of optimisation techniques spanning from the classical approaches such as nonlinear programming~\cite{khan2012optimization} to Evolutionary Algorithms (EA) such as Genetic Algorithms (GA) and Differential Evolution (DE)~\cite{beck2015multi,ibrahim2018optimization,faccenda1992combined,angira2003evolutionary}. Specifically, the popularity of EAs stems from them being sufficiently flexible across a broad range of problems. This flexibility arises from not requiring problem-specific analytical information such as differential equations and derivatives~\cite{rangaiah2009multi}. The framework for EAs start with an initial population (set of solutions) and converges these towards improved solution spaces using selection, crossover and mutation operators~\cite{back1996evolutionary}; with the literature showing promising results when applied to chemical process problems~\cite{woolway2019application,woolway2018novel,bowditch2019comparative,van2020makespan}.

One particular Chemical Process optimisation system, Pressure Swing Adsorption (PSA), has received a significant amount of interest~\cite{ye2019artificial,leperi2019110th,xiao2020machine,perez2019experimental}, with the typical PSA system usually involving optimisation across multiple objectives. A specific variant of the PSA system studied by Yancy-Cabellero \textit{et al.}~\cite{yancy2020process} has been evaluated in this study. This specific variant involves maximising two objectives, namely the CO$_{\text{2}}$ \textit{purity} and \textit{recovery}. Systems, such as PSA, with multiple objectives, usually have a set of ``Pareto-optimal" solutions, none of which are better than the others~\cite{deb2002fast}. EAs are common solutions in many multi-objective optimisation domains~\cite{jin2018data,jourdan2006lemmo}. NSGA-\Romannum{2} in particular is a popular EA optimisation approach that has been shown to be robust for the optimisation of a wide variety of these multiple-objective systems~\cite{pai2020experimentally,subraveti2019machine,haghpanah2013multiobjective,beck2015multi,haghpanah2013cycle}.

Combining complex simulations with EAs can result in higher computational costs as multiple evaluations of the simulation model are required to converge on an optimal solution. The existing literature on Surrogate Assisted optimisation has demonstrated that optimal solutions can be achieved with fewer function evaluations through substituting the complex simulation with a surrogate model~\cite{beykal2018optimal,chandra2022surrogate,chugh2017data,ibrahim2018optimization,park2018multi,perumal2020surrogate,van2021parden}. Here, a \textit{Surrogate Model} is a computationally cheaper model acting as a proxy for the original computationally costly simulation~\cite{razavi2012review,simpson2008design}. Surrogate models have been shown to have a wide range of applications in various domains. A few examples of the different areas include the optimisation of designing airfoils, turbine blades, vehicle crash tests, proteins, and drugs~\cite{jin2011surrogate,liu2022surrogate}. As the surrogate model is a substitute for the system under consideration, its selection is problem dependent and does not support a one-size-fits-all approach~\cite{kowsher2021support,damuluri2020analyzing}. The review paper completed by Bhoesekar \textit{et al.}~\cite{bhosekar2018advances} details that the surrogate model selection process is quite challenging and that problem categorisation into either feasibility, prediction or optimisation is beneficial to the selection of a suitable surrogate model.

Numerous techniques have been presented in the surrogate-assisted chemical engineering optimisation literature. Wahid \textit{et al.}~\cite{ali2018surrogate} and Shi \textit{et al.}~\cite{shi2016evolutionary} made use of Radial basis functions as their surrogate model for minimising compression energy in a single mixed refrigerant process of natural gas liquefaction and optimising crude oil distillation units. A Kriging surrogate model was implemented by Beck \textit{et al.}~\cite{beck2015multi} in optimising the design of a vacuum/pressure swing adsorption system. An Artificial Neural Network was implemented by Anna \textit{et al.}~\cite{anna2017machine} in optimising a pressure swing adsorption unit as well as by Ibrahim \textit{et al.}~\cite{ibrahim2018optimization} in the optimisation of the design of crude oil distillation units. Notably, in the EA literature, the use of a Surrogate Assisted Genetic Algorithm (SA-GA) has also been shown to speed up the optimisation of complex chemical systems while maintaining the quality of the solutions~\cite{art:stander2020,stander2020extended}.

Due to Surrogate Assisted optimisation's popularity and the massive potential benefits in the chemical engineering field, it is crucial to ensure that the existing surrogate assisted EA algorithms are robust and generalisable. Specifically, the existing solution~\cite{art:stander2020,stander2020extended} is not shown to be applicable to intricacies such as parallel systems and systems with feedback loops. Most poignantly, the previous approach has not been tested or validated on multi-objective extension or more complex real-world chemical systems such as PSA.

For this reason, we expand on the research by Stander \textit{et al.}~\cite{art:stander2020} to investigate the performance of the algorithms on complex higher-dimensional problems. To this end, this study includes further testing and validating of SA-GA and a multi-objective extension, Surrogate Assisted NSGA-\Romannum{2} (SA-NSGA), on more complex and real-world chemical systems.

The first set of systems this study extends to is two more complex variants of the example introduced by Stander \textit{et al.}~\cite{art:stander2020}. The additional complexities introduced include a parallel system and a feedback loop in the single-objective setting. The third chemical process investigated is the Pressure Swing Adsorption System (PSA). The PSA introduces a spectrum of complexities, including multiple feedback loops, to interrogate our novel multi-objective algorithm's robustness and generalisability.

The single-objective SA-GA extensions include testing various additional optimisation techniques selected to determine a significant difference in performance between the more complex and straightforward techniques. The optimisation techniques that have been implemented in this research include Nelder Mead (NM), which makes use of the simplex method to minimise functions~\cite{nelder1965simplex}. The second optimisation technique evaluated in this research is DE, an evolutionary algorithm similar to GAs better able to handle real number representations~\cite{fleetwood2004introduction,price2006differential}. The Improved Stochastic Ranking Evolutionary Strategy (ISRES) is the third optimisation technique evaluated. ISRES is an evolutionary strategy that balances the objective function and the constraints using a multi-objective formulation~\cite{runarsson2005search}. The final optimisation technique evaluated is the Particle Swarm Optimisation (PSO) which makes use of a particle set, called a swarm, to guide the search~\cite{pymoo,li2018two}.

For the multi-objective extension, this study evaluates NSGA-\Romannum{2} together with two additional multi-objective optimisation strategies for the PSA system. The additional strategies are the NSGA-\Romannum{3} technique and the C-TAEA technique; both algorithms are multi-objective EAs capable of handling constraints~\cite{pymoo}. NSGA-\Romannum{3} was developed as an improvement over the NSGA-\Romannum{2} for better handling of problems with numerous objectives~\cite{pymoo}, while the C-TAEA algorithm implements a two archive strategy to balance diversity and convergence~\cite{pymoo}. 

In addition to testing various optimisation techniques for the single and multi-objective systems, several Machine learning techniques, including the Decision Tree and Random Forest, were also evaluated. We include the tree-based machine learning techniques due to their ability to generalise across complex systems without requiring a large amount of training data, as supported by our empirical experimentation detailed below~\cite{talebkeikhah2021comparison}.

In previous work by Stander \textit{et al.}~\cite{art:stander2020} they show that the Evolutionary Algorithm implemented was able to optimise across a single objective stochastic system. The study also illustrated the computational time savings when substituting the simulation model with a surrogate Machine Learning model in combination with the evolutionary algorithm. We build upon that study with the following novel contributions that demonstrate that:
\begin{itemize}
    \item surrogate assisted Evolutionary Algorithms can be scaled to increasingly complex chemical systems such as systems with parallel and feedback components;
    \item Random Forest is a superior choice as a surrogate model across numerous plant design and optimisation problems;
    \item NSGA-\Romannum{2} outperforms the other meta-heuristic optimisation techniques and provides a sensible base model for surrogate improvements;
    \item this scaling applies to real-world systems such as a Pressure Swing Adsorption system with multi-objective optimisation requirements; and
    \item when meta-heuristic optimisation is combined with surrogate assisted techniques, the speedup is significant, and the results are both robust and extensible.
\end{itemize}

The chemical plant systems investigated in this study are presented in two separate sections. The first section includes two more complex variants of examples investigated by Stander \textit{et al.}~\cite{art:stander2020} referred to as the motivating examples. The second section contains the primary example in the form of a PSA system. The remainder of the paper includes the details of the chemical plant systems; the surrogate assisted optimisation methodology and the results for optimising each chemical plant system.

\section{Chemical Plant Systems}\label{sec:CPS}

This section details the Chemical Plant systems that were investigated. The first subsection includes the motivating examples; more complex variants, including a parallel system and a feedback loop system. The second subsection includes the primary example in the form of the PSA system.

\subsection{Motivating Examples - CPS Systems}\label{sec:plantDesign}

Two different Chemical Plant Systems (CPS) are illustrated in Figures ~\ref{Problem1cps}, \ref{Problem2cps}. The systems represented in these figures are referred to as CPS-1 (Figure \ref{Problem1cps}) and CPS-2 (Figure \ref{Problem2cps}). 

\tikzstyle{startstop} = [rectangle, rounded corners, minimum width=2cm, minimum height=1cm,text centered, draw=ashgrey, drop shadow, fill=ashgrey!50,]
\tikzstyle{io} = [trapezium, trapezium left angle=70, trapezium right angle=110, minimum width=2cm, minimum height=1cm, text centered, draw=black, drop shadow]
\tikzstyle{process} = [rectangle, minimum width=2cm, minimum height=1cm, text centered, text width=3cm, draw=orange, drop shadow, fill = orange!20]
\tikzstyle{process2} = [rectangle, minimum width=2cm, minimum height=1cm, text centered, text width=2cm, draw=black, drop shadow, fill = white]
\tikzstyle{decision} = [diamond, minimum width=3cm, minimum height=1cm, text centered, draw=darkred, drop shadow, fill = darkred!20]
\tikzstyle{arrow} = [thick,->,>=stealth]
\tikzset{Tank/.style={draw=black,fill=white,thick,rectangle,rounded corners=10pt,minimum width=0.75cm,minimum height=1.5cm,text width=0.75cm,align=center, drop shadow}}

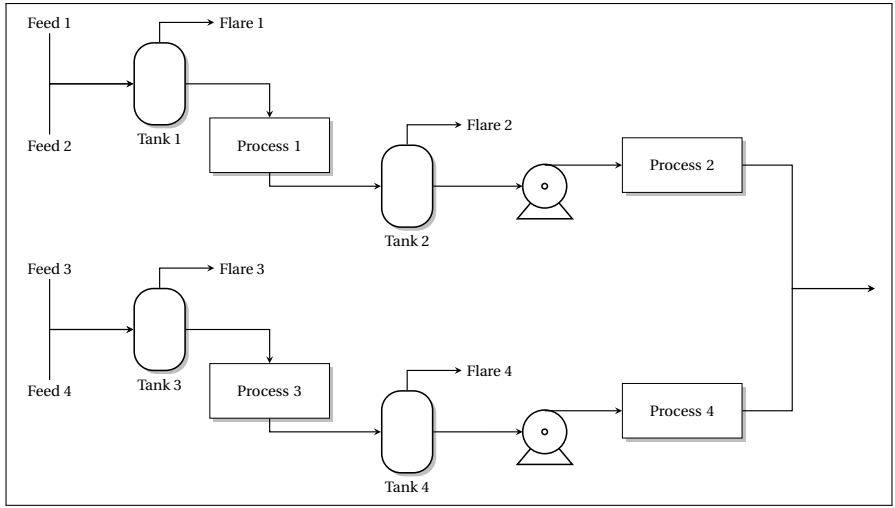
\begin{figure*}
    \centering
    \fbox{
    \resizebox{.65\textwidth}{!}{
    \begin{tikzpicture}[yscale=0.75]
    \node (feed1) at (0, 0) {Feed 1};
    \node (feed2) at (0, -3) {Feed 2};
    \node (flare1) at (3.5, 0) {Flare 1};
    \node (flare2) at (8, -2.5) {Flare 2};
    \node[Tank] (t1) at (2, -1.5) {};
    \node[below of = t1] {Tank 1};
    \node[process2] (proc1) at (4, -3) {Process 1};
    \node[Tank] (t2) at (6.5, -4) {};
    \node[below of = t2] {Tank 2};
    \pic[scale = 1] at (9, -4) {centrifugal pump};
    \node[process2] (proc2) at (11.5, -3.485) {Process 2};
    
    \node (feed3) at (0, -6) {Feed 3};
    \node (feed4) at (0, -9) {Feed 4};
    \node (flare3) at (3.5, -6) {Flare 3};
    \node (flare4) at (8, -8.5) {Flare 4};
    \node[Tank] (t3) at (2, -7.5) {};
    \node[below of = t3] {Tank 3};
    \node[process2] (proc3) at (4, -9) {Process 3};
    \node[Tank] (t4) at (6.5, -10) {};
    \node[below of = t4] {Tank 4};
    \pic[scale = 1] at (9, -10) {centrifugal pump};
    \node[process2] (proc4) at (11.5, -9.485) {Process 4};
    \node (finalproduct) at (14.2, -6.2) {};
   
   \draw[main stream] (feed1) |- (t1);
   \draw[main stream] (feed2) |- (t1);
   \draw[main stream] (t1) |- (flare1);
   \draw[main stream] (t1) -| (proc1);
   \draw[main stream] (proc1) |- (t2);
   \draw[main stream] (t2) |- (flare2);
   \draw[main stream] (t2) -- (8.6, -4);
   \draw[main stream] (9, -3.485) -- (proc2);
   \draw[main stream] (proc2) -- (13.5, -3.485) |- (15, -6.5);
   
   \draw[main stream] (feed3) |- (t3);
   \draw[main stream] (feed4) |- (t3);
   \draw[main stream] (t3) |- (flare3);
   \draw[main stream] (t3) -| (proc3);
   \draw[main stream] (proc3) |- (t4);
   \draw[main stream] (t4) |- (flare4);
   \draw[main stream] (t4) -- (8.6, -10);
   \draw[main stream] (9, -9.485) -- (proc4);
   \draw[main stream] (proc4) -- (13.5, -9.485) |- (15, -6.5);
   \end{tikzpicture}
   }
   }
   \caption{CPS-1 - A Chemical Plant System with Parallel Streams}
\label{Problem1cps}
\end{figure*}

\begin{figure*}
    \centering
    \fbox{
    \resizebox{.65\textwidth}{!}{
    \begin{tikzpicture}[yscale=0.75]
    \node (feed1) at (0, 0) {Feed 1};
    \node (feed2) at (0, -3) {Feed 2};
    \node (flare1) at (3.5, 0) {Flare 1};
    \node (flare2) at (8, -2.5) {Flare 2};
    \node[Tank] (t1) at (2, -1.5) {};
    \node[below of = t1] {Tank 1};
    \node[process2] (proc1) at (4, -3) {Process 1};
    \node[Tank] (t2) at (6.5, -4) {};
    \node[below of = t2] {Tank 2};
    \pic[scale = 1] at (9, -4) {centrifugal pump};
    \node[process2] (proc2) at (11.5, -3.485) {Process 2};
    \node (Feedback) at (7,0.8) {\begin{math}\alpha\end{math}};
    \node (feedbackper) at (13.25, -3.1) {1 - \begin{math}\alpha\end{math}};
   \draw[main stream] (feed1) |- (t1);
   \draw[main stream] (feed2) |- (t1);
   \draw[main stream] (t1) |- (flare1);
   \draw[main stream] (t1) -| (proc1);
   \draw[main stream] (proc1) |- (t2);
   \draw[main stream] (t2) |- (flare2);
   \draw[main stream] (t2) -- (8.6, -4);
   \draw[main stream] (9, -3.485)  -- (proc2);
   \draw[main stream] (proc2) -- (14, -3.485);
   \draw[main stream] (11.5,-2.8) -- (11.5,0.5)-- (1.75,0.5) -- (1.75, -0.55);
   \end{tikzpicture}
   }
   }
   \caption{CPS-2 - A Chemical Plant System with Feedback Loop}
\label{Problem2cps}
\end{figure*}
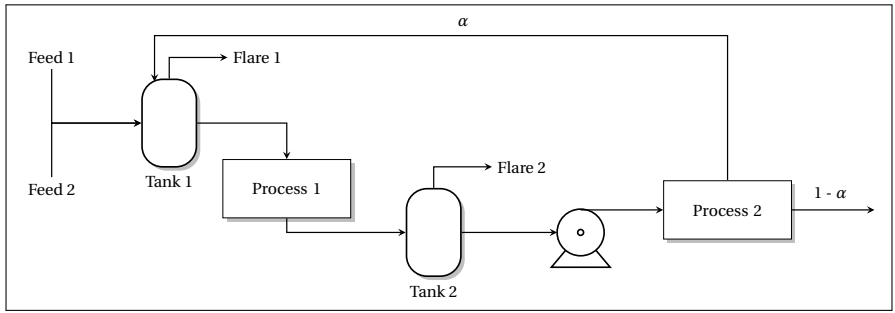

CPS-1 (Figure \ref{Problem1cps}) differs from the original system due to introducing a parallel system. The parallel system adds complexity by having two stochastic systems relying on a centralised pool of spares for both systems' failures. This shared pool needs to be managed to reduce the failure times for both systems and the total cost associated with the number of spares procured. The system's decision variables include tank capacities, pump capacities, the minimum number of spares to keep on-site, and the minimum number of pumps to procure. The introduced specification changes include doubling the maximum bounds of the number of spares to keep and the number of spares to purchase and introducing three more decision variables that included two tanks and pump capacities.

CPS-2 (Figure \ref{Problem2cps}) adds the complexity of a recycle stream: product that has not met quality specifications and needs to be reprocessed. The ratio of product recycled to the final product is represented by $\alpha$. The critical lever to influencing this ratio is the number of maintenance hours spent on the running system. The $\alpha$ ratio is sampled from a normal distribution with a mean that is a function of the number of maintenance hours spent and has a variance of 10\% of the mean maintenance hours. The optimiser can select the number of maintenance hours between 0 and 1314, where the maximum bound represents 15\% of the entire cycle time of the simulation model. The rest of the decision variables include tank capacities, pump capacity, the minimum number of spares to keep on-site and the minimum number of pumps to procure.

The objective value for both motivating examples is maximising the overall revenue of the system. The main revenue generator is the final product produced. The systems' revenues are impacted negatively by the cost of flaring product and failures on the pumps and processing units. The objective functions for the two systems are similar to the details summarised in the paper by Stander \textit{et al.}~\cite{art:stander2020}, focusing on maximising the final product stream and minimising all the other variables. Table \ref{tab5} includes the costs and revenues for the fitness function values for the two systems and is defined as:
\begin{equation}
\operatorname*{argmax}_{x_1} \quad  c_1 x_1 - \sum_{i=2}^n c_i x_i,
\end{equation}
where $n$ is the number of costs to include for the respective systems.

\begin{table}[htbp]
\caption{Fitness Function Values}
\centering
\resizebox{.975\columnwidth}{!}{
\begin{tabular}{ rlclc } 
\toprule
{\bm{$(i)$}} & {\textbf{Equipment/Process} \bm{$(x_i)$}} & {\textbf{Revenue/Costs} \bm{$(c_i)$}} & {\textbf{CPS Set}} & {\textbf{Bounds}}\\
\bottomrule\toprule
1 & Final Product & 6042/m$^{\mathrm{3}}$ & Both & -\\ 
2 & Flare 1 & 2848/m$^{\mathrm{3}}$ & Both & -\\ 
3 & Flare 2 & 3907/m$^{\mathrm{3}}$ & Both & -\\  
4 & Per Tank   & \num{9.94e7} $+$ \num{1.52e6}/m$^{\mathrm{3}}$ & Both & [500, 1000]\\
5 & Per Pump   & \num{4.44e6} $+$ \num{2.96e5}/m$^{\mathrm{3}}$ & Both & [60, 120]\\
6 & Flare 3 & 2848/m$^{\mathrm{3}}$ & CPS-1 & -\\ 
7 & Flare 4 & 3907/m$^{\mathrm{3}}$ & CPS-1 & -\\
8 & Number of Man Hours & 474036/hr & CPS-2 & [0, 1314]\\
\bottomrule
\end{tabular}
}
\label{tab5}
\end{table}

\subsection{Primary Example - PSA System}
\label{sec:PSA}

A post-combustion CO$_{\text{2}}$ capture PSA process has been used to further test and validate SA-GA. The PSA system was sourced from work completed by Yancy-Cabellero \textit{et al.}~\cite{yancy2020process}. The carbon capture process plays a significant role in reducing the CO$_{\text{2}}$ emissions from gas-fired and coal power plants~\cite{yancy2020process}. The PSA system forms part of the first step of the three-step carbon capture process~\cite{yancy2020process}.

The PSA cycle is illustrated in Figure \ref{fig:PSA_Cycles}. The PSA cycle configuration that has been implemented in this study is called the modified Skarstrom cycle. The system consists of five steps within a cycle, namely: A - Pressurisation, B - Adsorption, C - Heavy Reflux, D - Counter-Current Depressurisation and E - Light Reflux~\cite{yancy2020process}. The cycle starts at low pressure and is then pressurised (step A) by the flue gas up to the adsorption pressure. Once the adsorption pressure is reached, the Adsorption step (step B) takes place by having the top end of the column open, allowing the feed gas to be fed through and the CO$_{\text{2}}$ to be concentrated at the opposite end of the column. The Heavy Reflux step involves substituting the flue gas flowing into the column by the heavy product collected during the Light Reflux step (step E) after a set amount of time. This substitution results in a higher concentration of CO$_{\text{2}}$ at the bottom of the column due to the heavy product's high CO$_{\text{2}}$ concentration. The bottom of the column is then closed, and the pressure in that section is dropped to the starting pressure during the Counter-Current Depressurisation step (step D). This step also includes collecting emissions at the bottom of the column as the CO$_{\text{2}}$ product. The final step, the Light Reflux step (step E), involves feeding the light product produced in the adsorption phase into the top end of the column once the initial pressure is achieved, and the cycle repeats~\cite{yancy2020process}.

\tikzstyle{startstop} = [rectangle, rounded corners, minimum width=2cm, minimum height=1cm,text centered, draw=ashgrey, drop shadow, fill=ashgrey!50,]
\tikzstyle{io} = [trapezium, trapezium left angle=70, trapezium right angle=110, minimum width=2cm, minimum height=1cm, text centered, draw=black, drop shadow]
\tikzstyle{process} = [rectangle, minimum width=2cm, minimum height=1cm, text centered, text width=3cm, draw=orange, drop shadow, fill = orange!20]
\tikzstyle{process2} = [rectangle, minimum width=2cm, minimum height=1cm, text centered, text width=2cm, draw=black, drop shadow, fill = white]
\tikzstyle{decision} = [diamond, minimum width=3cm, minimum height=1cm, text centered, draw=darkred, drop shadow, fill = darkred!20]
\tikzstyle{arrow} = [thick,->,>=stealth]
\tikzset{Tank/.style={draw=black,fill=white,thick,rectangle,rounded corners=10pt,minimum width=0.75cm,minimum height=1.5cm,text width=0.75cm,align=center, drop shadow}}

\begin{figure*}
    \centering
    \fbox{
    \resizebox{.65\textwidth}{!}{
    \begin{tikzpicture}[yscale=0.75]
    \node [Tank](t1) at (0, 0) {};
     \node[Tank] (t2) at (1.5, 0) {};
     \node[Tank] (t3) at (3, 0) {};
     \node[Tank] (t4) at (4.5, 0) {};
     \node[Tank] (t5) at (6, 0) {};

     \draw[main stream](1.5, 1) -- (1.5,2);
     \draw[main stream](3, 1) -- (3,2);
     \draw[main stream](1.5, 1.5) -- (6, 1.5) -| (6,1);
     \draw[main stream](0.75, -1.5) -- (0,-1.5) -| (0,-1);
     \draw[main stream](0.75, -1.5) -- (1.5,-1.5) -| (1.5,-1);
     \draw[main stream](-1.5, -2) -- (0.75,-2) -| (0.75,-1.5);
     \draw[main stream](4.5, -1) -- (4.5,-2) |- (7,-2);
     \draw[main stream](6, -1) -- (6,-2);
     \draw[main stream](6, -1.5) -- (3,-1.5) -| (3,-1);

     \node at (1.5,2.25) {N$_\textit{2}$};
     \node at (3,2.25) {N$_\textit{2}$};
     \node at (-2.3, -1.5) {Flue Gas};
     \node at (-2.3, -2) {15$\%$ CO$_\textit{2}$};
     \node at (-2.375, -2.45) {85$\%$ N$_\textit{2}$};
     \node at (7.5, -2) {CO$_\textit{2}$};
     \node at (0, 0) {A};
     \node at (1.5, 0) {B};
     \node at (3, 0) {C};
     \node at (4.5, 0) {D};
     \node at (6, 0) {E};
     
  \end{tikzpicture}
  }
  }
  \caption{PSA System - A 5 step modified Skarstrom cycle (A - Pressurisation, B - Adsorption, C - Heavy Reflux, D - Counter-Current Depressurisation and E - Light Reflux) \cite{yancy2020process}}
\label{fig:PSA_Cycles}
\end{figure*}
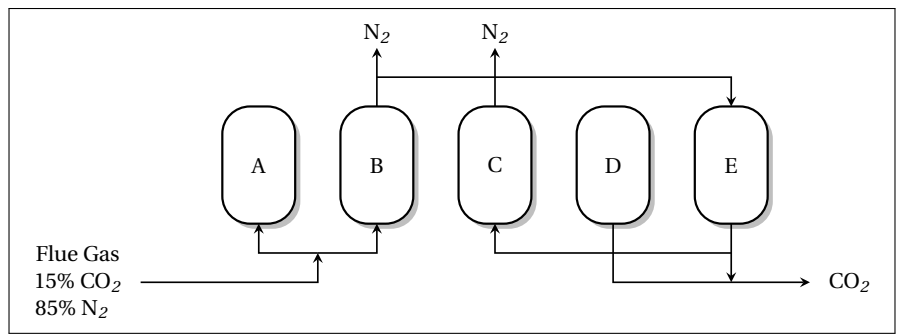

The PSA system's simulation model was developed in \textsc{Matlab} from a set of partial differential equations describing the system. To optimise the PSA system, the two operating parameters that are maximised at each cycle are CO$_{\text{2}}$:
\begin{equation}
\label{eq:Purity}
\textit{Purity} = \frac{\mbox{Moles of $CO_2$ in product}}{\mbox{Total moles in the Product}} \times 100\%
\end{equation}
and:
\begin{equation}
\label{eq:Recovery}
\textit{Recovery} = \frac{\mbox{Moles of $CO_2$ in to product}}{\mbox{Moles of $CO_2$ fed into cycle}} \times 100\%.
\end{equation}

\section{Methodology}
\label{sec:methodology}

The methodology section details the surrogate assisted optimisation algorithm and its various components for the motivating and primary examples. Each subsection detail the algorithm implemented, the sorting mechanism implemented, the crossover and mutation used, the surrogate models, and each chromosome's gene makeup within the population. The PSA chemical system section includes two additional subsections detailing the additional metrics used to analyse the various experiments tested and the details of the software and packages used for the different implementations.

\subsection{Motivating Examples - CPS Systems}

The additional contribution for these two systems focuses on testing the SA-GA on the more complex motivating example variants. The changes introduced into the methodology related to the optimisation technique, where the chromosome's gene structure was changed to include the additional decision variables for each system.

The motivating examples were simulated using stochastic continuous-time event simulation for $365$ days in hourly time increments, evaluating the objective functions per simulation run for both systems. These parameters were selected for the simulation as a year captures all possible events that could occur. The hourly increments represent the smallest increment required to illustrate changes in the system.

\subsubsection{Meta-heuristic Optimisation Technique Selection}

\begin{table}[htbp]
\caption{Optimisation Technique Comparison - CPS System Revenue Results}
\centering
\begin{tabular}{ llr } 
\toprule
    \textbf{System} &
    \textbf{Opt' Technique}  & 
    \textbf{Max Revenue}   \\
\bottomrule
\toprule
\multirow{5}{*}{\makecell[lt]{CPS 1}}
& GA                 & $\mathbf{3.1 \pm 0.138 \times 10^{9}}$\\  
& DE                 & $3.0 \pm 0.188 \times 10^{9}$ \\  
& ISRES              & $3.0 \pm 0.157 \times 10^{9}$ \\ 
& PSO                & $3.0 \pm 0.124 \times 10^{9}$ \\
& NM                 & $1.8 \pm 0.421 \times 10^{9}$ \\ 
\midrule
\multirow{5}{*}{\makecell[lt]{CPS 2}}
& GA                 & $\mathbf{1.2 \pm 0.085 \times 10^{9}}$ \\  
& DE                 & $\mathbf{1.2 \pm 0.104 \times 10^{9}}$ \\  
& ISRES              & $1.1 \pm 0.109 \times 10^{9}$ \\ 
& PSO                & $\mathbf{1.2 \pm 0.010 \times 10^{9}}$ \\
& NM                 & $2.3 \pm 2.816 \times 10^{8}$ \\ 
\bottomrule
\label{tab:opt-tech-comp-cps}
\end{tabular}
\end{table}

The CPS systems were optimised using a range of meta-heuristic strategies using out of the box hyper-parameters to determine which performs best. This result helped determine which meta-heuristic strategy will be combined with a surrogate for further validation of our proposed method. The aggregated results of thirty runs of each of these experiments are summarised in Table~\ref{tab:opt-tech-comp-cps}. The meta-heuristic optimisation algorithms implemented are GAs, DE, ISRES and PSO. We also provide the results for NM as a baseline for comparison. These algorithms were selected to test a range of simple to more complex techniques and determine the performance differences. The best performing algorithm across the CPS systems was the GA and was selected for further investigation to optimise the CPS systems. Implementation details for the Surrogate Assisted GA are expanded further in the following section.

\subsubsection{Surrogate Assisted GA Algorithm}\label{algorithm_cps}

The Surrogate Assisted Genetic Algorithms (SA-GA) utilised for the motivating examples starts with an initial random population of size 800. The best 75 candidates of the initial population are selected as a warm start population. The surrogate machine learning model (Random Forest) is initialised using this warm start population.

The SA-GA continues from here following the flow-sheet steps in Figure \ref{algo}. The initial population is ranked from highest to lowest revenue, enabling the top 15\% of the population (elite) to be selected. This elite is used to generate offspring via crossover using the BLX-$\alpha$ algorithm~\cite{eshelman1993real}. Mutation via random substitution is then applied at a rate of 30\%. 

To manage deviations between their predictions: at each generation, the elite population is evaluated by both the surrogate and the simulation model. The deviation is assessed by calculating the difference between the surrogate predictions and the simulation model's output. If the difference is greater than 1$\sigma$, then the simulation model is used as the evaluation platform, and the surrogate model is retrained with the new data generated by the simulation. The algorithm is run for fifty generations before terminating. The SA-GA parameters were determined through the testing phase of previous work by Stander \textit{et al.}~\cite{art:stander2020,stander2020extended}. A range of starting population sizes, elite percentages, crossover techniques, mutation rates and maximum generations were tested to determine the best performing parameters.

\tikzstyle{startstop} = [rectangle, rounded corners, minimum width=2cm, minimum height=1cm,text centered, draw=ashgrey, drop shadow, fill=ashgrey!50,]
\tikzstyle{io} = [trapezium, trapezium left angle=70, trapezium right angle=110, minimum width=2cm, minimum height=1cm, text centered, draw=black, drop shadow]
\tikzstyle{process} = [rectangle, minimum width=2cm, minimum height=1cm, text centered, text width=3cm, draw=orange, drop shadow, fill = orange!20]
\tikzstyle{process2} = [rectangle, minimum width=2cm, minimum height=1cm, text centered, text width=2cm, draw=black, drop shadow, fill = white]
\tikzstyle{decision} = [diamond, minimum width=3cm, minimum height=1cm, text centered, draw=darkred, drop shadow, fill = darkred!20]
\tikzstyle{arrow} = [thick,->,>=stealth]
\tikzset{Tank/.style={draw=black,fill=white,thick,rectangle,rounded corners=10pt,minimum width=0.75cm,minimum height=1.5cm,text width=0.75cm,align=center, drop shadow}}

\tikzstyle{line} = [draw, -stealth]
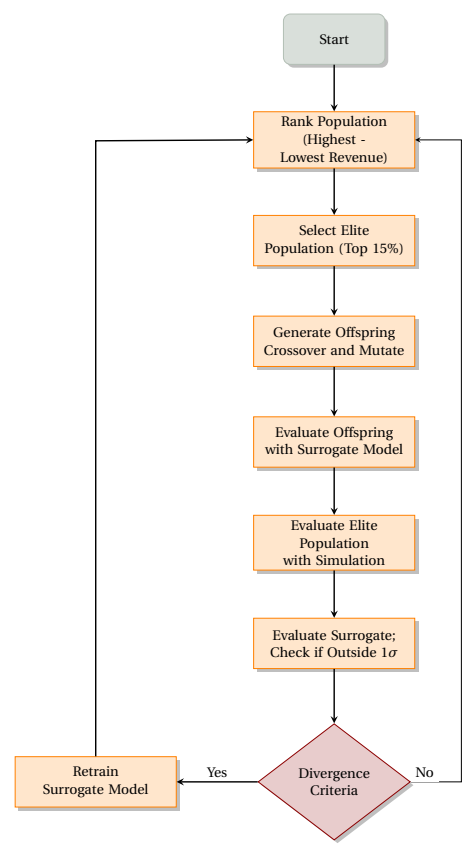
\begin{figure}[htbp]
\centering
\resizebox{0.43\textwidth}{!}{
\begin{tikzpicture}[node distance=1cm,every node/.style={fill=white}, align=center]
\node (start) [startstop,xshift=3cm] {Start};
\node (pro1) [process, below of =start, yshift=-1cm] {Rank Population (Highest - Lowest Revenue)};
\node (pro2) [process, below of =pro1, yshift=-1cm] {Select Elite Population (Top 15\%)};
\node (pro3) [process, below of =pro2, yshift=-1cm] {Generate Offspring Crossover and Mutate};
\node (pro4) [process, below of =pro3, yshift=-1cm] {Evaluate Offspring with Surrogate Model};
\node (pro5) [process, below of =pro4, yshift=-1cm] {Evaluate Elite Population with Simulation};
\node (pro6) [process, below of=pro5,  yshift=-1cm] {Evaluate Surrogate; Check if Outside $1\sigma$};
\node (dec2) [decision, below of=pro6, yshift=-1.75cm] {Divergence\\Criteria};
\node (pro7) [process, left of =dec2,  xshift=-3.7cm] {Retrain Surrogate Model};
\node (inv1) [ right of=dec2, xshift=3.7cm] {};
\draw [arrow] (start) -- (pro1);
\draw [arrow] (pro1) -- (pro2);
\draw [arrow] (pro2) -- (pro3);
\draw [arrow] (pro3) -- (pro4);
\draw [arrow] (pro4) -- (pro5);
\draw [arrow] (pro5) -- (pro6);
\draw [arrow] (pro6) -- (dec2);
\draw [arrow] (dec2) -- node[above]{Yes}(pro7);
\path [line] (dec2.east) node [anchor =south west] {No} --+(1cm,0)  |-(pro1);
\draw[arrow] (pro7.north) -- ++(0,5.35) -- ++(0,6.875) --(pro1.west);
\end{tikzpicture}
}
\caption{Flow Chart for Surrogate Assisted Optimisation Algorithm using Genetic Algorithms}
\label{algo}
\end{figure}

\subsubsection{Crossover and Mutation} \label{crossover_mutation_cps}

\begin{figure}[htbp]
\centering
\begin{tikzpicture}[node distance=1cm,align=center, scale = 1.25]
\draw[fill=gray!30, drop shadow] (0,1) rectangle (0.5,1.5) node [xshift = -9, yshift = -8] {\footnotesize 600};
\draw[fill=gray!30, drop shadow] (0.5,1) rectangle (1,1.5) node [xshift = -9, yshift = -8] {\footnotesize 750};
\draw[fill=white,drop shadow] (1,1) rectangle (1.5,1.5) node [xshift = -9, yshift = -8] {\footnotesize 80};
\draw[fill=white,drop shadow] (1.5,1) rectangle (2,1.5) node [xshift = -9, yshift = -8] {\footnotesize 5};
\draw[fill=white,drop shadow] (2,1) rectangle (2.5,1.5) node [xshift = -9, yshift = -8] {\footnotesize 8};
\draw[fill=white,drop shadow] (2.5,1) rectangle (3,1.5) node [xshift = -9, yshift = -8] {\footnotesize 650};
\node [left] at (0,1.25 ) {Chromosome 2};
\draw[fill=white,drop shadow] (0,2) rectangle (0.5,2.5) node [xshift = -9, yshift = -8] {\footnotesize 820};
\draw[fill=white,drop shadow] (0.5,2) rectangle (1,2.5) node [xshift = -9, yshift = -8] {\footnotesize 900};
\draw[fill=gray!30,drop shadow] (1,2) rectangle (1.5,2.5) node [xshift = -9, yshift = -8] {\footnotesize 110};
\draw[fill=gray!30,drop shadow] (1.5,2) rectangle (2,2.5) node [xshift = -9, yshift = -8] {\footnotesize 2};
\draw[fill=gray!30,drop shadow] (2,2) rectangle (2.5,2.5) node [xshift = -9, yshift = -8] {\footnotesize 4};
\draw[fill=gray!30,drop shadow] (2.5,2) rectangle (3,2.5) node [xshift = -9, yshift = -8] {\footnotesize 300};
\node [left] at (0,2.25 ) {Chromosome 1};
\node[fill=ashgrey!50, style={single arrow, draw=none, rotate=270}] at (1., 0.125) {crossover};
\draw[fill=white,drop shadow] (0,-1.25) rectangle (0.5,-.75) node [xshift = -9, yshift = -8] {\footnotesize 3};
\draw[fill=white,drop shadow] (0.5,-1.25) rectangle (1,-.75) node [xshift = -9., yshift = -8] {\footnotesize 900};
\draw[fill=white,drop shadow] (1,-1.25) rectangle (1.5,-.75) node [xshift = -9, yshift = -8] {\footnotesize 0.9};
\draw[fill=white,drop shadow] (1.5,-1.25) rectangle (2,-.75) node [xshift = -9, yshift = -8] {\footnotesize 1.8};
\draw[fill=white,drop shadow] (2,-1.25) rectangle (2.5,-.75) node [xshift = -9, yshift = -8] {\footnotesize 0.85};
\draw[fill=white,drop shadow] (2.5,-1.25) rectangle (3,-.75) node [xshift = -9, yshift = -8] {\footnotesize 0.2};
\node [left] at (0,-1. ) {Child 1};

\draw [darkred] (0.9, 2.65) rectangle (1.1, 0.9);

\end{tikzpicture}
\caption{An Example of the Crossover Operation for two of the potential candidates used by the GA}
\label{fig:crossover_cps2}
\end{figure}
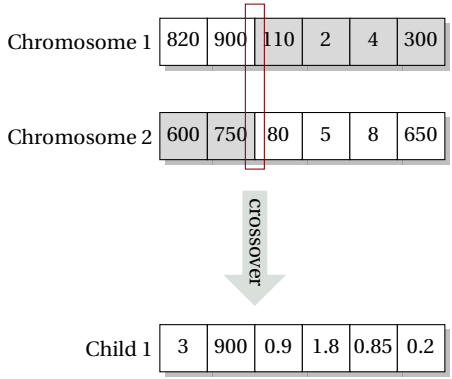

The BLX-$\alpha$ crossover technique involves combining a pair of chromosomes ~\cite{eshelman1993real}. The BLX-$\alpha$ algorithm works in two steps for each gene in a chromosome:
\begin{enumerate}
    \item find the minimum ($\text{min}$) and the maximum ($\text{max}$) values of the parent genes and calculate the $\text{range} = \text{max} - \text{min}$,
    \item the child gene will be a random number in the interval $[\text{min} - (\text{range} \times \alpha), \text{max} + (\text{range} \times \alpha)]$,
\end{enumerate}
where $\alpha$ controls how much outside the $[\text{min}$,$\text{max}]$ interval you would like to consider. A value of $\alpha=0$ gives Uniform Crossover. A value of $\alpha=0.15$ is employed for this study previously shown to give good results~\cite{art:stander2020}. Every iteration, each new child's fitness value is compared to the fitness values of the elite population, and if it is discovered to be better, the chromosome is substituted into the elite population, and the cycle continues \cite{chudasama2011comparison}. Figure \ref{fig:crossover_cps2} is an example of the crossover procedure for CPS-2 illustrated in Figure \ref{Problem2cps}.

\begin{figure}[htbp]
\centering
\begin{tikzpicture}[node distance=1cm,align=center, scale = 1.25]
\draw[fill=white,drop shadow] (0,1) rectangle (0.5,1.5) node [xshift = -9, yshift = -8] {\footnotesize 7};
\draw[fill=white,drop shadow] (0.5,1) rectangle (1,1.5) node [xshift = -9, yshift = -8] {\footnotesize 20};
\draw[fill=gray!30, drop shadow] (1,1) rectangle (1.5,1.5) node [xshift = -9, yshift = -8] {\footnotesize 0.75};
\draw[fill=white,drop shadow] (1.5,1) rectangle (2,1.5) node [xshift = -9, yshift = -8] {\footnotesize 1};
\draw[fill=white,drop shadow] (2,1) rectangle (2.5,1.5) node [xshift = -9, yshift = -8] {\footnotesize 0.8};
\draw[fill=white,drop shadow] (2.5,1) rectangle (3,1.5) node [xshift = -9, yshift = -8] {\footnotesize 0.4};
\node [left] at (0,1.25 ) {Child 1};
\node[fill=ashgrey!50, style={single arrow, draw=none, rotate=270}] at (1.3, 0.125) {mutation};
\draw[fill=white,drop shadow] (0,-1.45) rectangle (0.5,-.95) node [xshift = -9, yshift = -8] {\footnotesize 7};
\draw[fill=white,drop shadow] (0.5,-1.45) rectangle (1,-.95) node [xshift = -9., yshift = -8] {\footnotesize 20};
\draw[fill=white,drop shadow] (1,-1.45) rectangle (1.5,-.95) node [xshift = -9, yshift = -8] {\footnotesize 0.5};
\draw[fill=white,drop shadow] (1.5,-1.45) rectangle (2,-.95) node [xshift = -9, yshift = -8] {\footnotesize 1};
\draw[fill=white,drop shadow] (2,-1.45) rectangle (2.5,-.95) node [xshift = -9, yshift = -8] {\footnotesize 0.8};
\draw[fill=white,drop shadow] (2.5,-1.45) rectangle (3,-.95) node [xshift = -9, yshift = -8] {\footnotesize 0.4};
\node [left] at (0,-1.2 ) {Child 1 Mutated};

\draw [darkred] (0.9, 1.6) rectangle (1.6, 0.9);

\end{tikzpicture}
\caption{An Example of the Mutation Operation for the potential candidate used by the GA}
\label{fig:mutation_cps}
\end{figure}
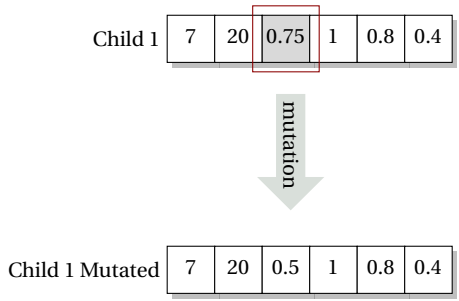

The crossover rate used for the optimisation was 100\%, meaning that the entire offspring was created from the elite population. The final step was the mutation of the offspring. Mutation was executed through random substitution with a random number within the bounds of that gene. Figure \ref{fig:mutation_cps} is an example of the mutation procedure.

\subsubsection{Genetic Structure} \label{psa_genetic_structure}
\begin{table}[htbp]
\caption{Genetic Structure for Gene 1 to 9}
\centering
\resizebox{.975\columnwidth}{!}{
\begin{tabular}{ rllc } 
\toprule
{\textbf{\#}} & {\textbf{Description}} & {\textbf{CPS Set}}& {\textbf{Bounds}}\\
\bottomrule\toprule
1  & Tank 1 Size  & Both & [500,1000]\\ 
2 & Tank 2 Size  & Both & [500,1000]\\ 
3 & Tank 3 Size  & CPS-1 & [500,1000]\\ 
4 & Tank 4 Size  & CPS-1 & [500,1000]\\ 
5 & Pump 1 Size  & Both & [60,120]\\
6 & Pump 2 Size  & CPS-1 & [60,120] \\
7 & Minimum Spares Level  & Both & [0,20]\\ 
8 & Minimum Spares to Purchase  & Both & [1,20]\\
9 & Number of Maintenance Man Hours  & CPS-2 & [0, 1314] \\
\bottomrule
\end{tabular}
}
\label{prob1gene}
\end{table}

The decision variables detailed in Section \ref{sec:CPS} formed part of the chromosomes that made up the GA population. The chromosomes' genetic structures used in CPS-1 and CPS-2 are summarised in Table \ref{prob1gene}. The genes were encoded with real numbers within the bounds specified in Table \ref{prob1gene}.

\subsubsection{Surrogate Model Selection} \label{surrogate_cps}

\begin{table}[htbp]
\caption{Machine Learning Surrogate Comparison - CPS System Revenue Results Relative to GA}
\centering
\begin{tabular}{ llr } 
\toprule
    \textbf{System} &
    \textbf{ML Surrogate}  & 
    \textbf{\makecell[r]{Revenue Change (\%)}}  \\
\bottomrule
\toprule
\multirow{6}{*}{\makecell[lt]{CPS 1}}
& Random Forest      & $\mathbf{1.04}$ \\  
& Decision Tree      & $\mathbf{1.04}$ \\ 
& Linear Regression  & $0.81$ \\ 
& Bayes Regression   & $0.73$ \\
& MLP                & $0.35$ \\
& Ridge Regression   & $0.65$ \\\midrule
\multirow{6}{*}{\makecell[lt]{CPS 2}}
& Random Forest      & $\mathbf{0.90}$ \\  
& Decision Tree      & $\mathbf{1.21}$ \\ 
& Linear Regression  & $0.78$ \\ 
& Bayes Regression   & $0.75$ \\
& MLP                & $0.25$ \\
& Ridge Regression   & $0.77$ \\
\bottomrule
\label{tab:surr-perf-comp-cps}
\end{tabular}
\end{table}

\begin{table}[htbp]
\caption{Hyper-parameter Set}
\centering
\resizebox{.975\columnwidth}{!}{
\begin{tabular}{ rlc } 
\toprule
{\textbf{\#}} & \textbf{Hyper-parameter} & {\textbf{Value}}\\
\bottomrule\toprule
1  & Number of Trees & 100\\ 
2 & Criterion & Mean Squared Error \\ 
3 & Splitting Minimum Samples & 2 \\
4 & Leaf Node Minimum Samples& 1 \\ 
5 & Splitting Maximum Features & 5 \\
6 & Bootstrapping & True \\
\bottomrule
\end{tabular}
}
\label{tab8}
\end{table}

Several Machine learning techniques were tested to determine the best performing algorithm for use as a surrogate model. The Random Forest algorithm was selected as the surrogate model due to it previously outperforming other tested algorithms~\cite{art:stander2020} as verified in the results summarised in Table~\ref{tab:surr-perf-comp-cps}. The hyper-parameters for the Random Forest are summarised in Table~\ref{tab8}.

\subsection{Primary Example - PSA System}

The methodology for the PSA system is a combination of the NSGA-\Romannum{2} strategy and the SA-GA algorithm~\cite{art:stander2020,stander2020extended}. This approach is aimed at firstly demonstrating that similar results can be achieved by replicating the existing implementation done by Yancy-Cabellero \textit{et al.}~\cite{yancy2020process}. It is secondly highlighting the computational cost reduction from using surrogate models as substitutes for the simulation model.

\subsubsection{Performance Metrics} \label{psa-perf}

The complexity of multi-objective optimisation necessitates additional performance metrics. Riquelme \textit{et al.}~\cite{riquelme2015performance} investigated and reviewed numerous performance metrics for multi-objective optimisation and concluded that the most popular metrics include the Hypervolume (HV), Generational Distance, Inverted Generational Distance and Epsilon Indicator~\cite{riquelme2015performance}. The HV metric is an area type calculations that is independent of any ideal solution. The GD and IGD metrics are distance measures to some reference vector. We adopt HV, Generational Distance Plus (GD+) and Inverted Generational Distance Plus (IGD+). The GD+ and IGD+ represent slightly modified versions of the original distance metrics taking into account the dominance of a solution \cite{ishibuchi2015modified,fonseca2006improved}. These two metrics are more robust to not having the ground truth Pareto front. We also use visual inspection of the Pareto frontiers, a common approach, in this study. 

The GD+ metric is represented by the following formula:
\begin{equation}
\label{eq:gd}
GD(X)^{+} = \frac{1}{\lvert X\rvert}\left(\sum_{i=1}^{\lvert X\lvert} {d_i^{+}}^{2}\right)^{\frac{1}{2}},
\end{equation}
where $X$ represents the solution vector achieved by the algorithm. The Euclidean distance from $x_i$ to the nearest reference point in the Pareto front or reference set of points is represented by ${d_i}^{+}$. The formula determines the average distance from any point on $X$ to the closest point on the (ideal) Pareto front.

The IGD+ formula is represented as:
\begin{equation}
\label{eq:igd}
IGD(X)^{+} = \frac{1}{\lvert Z\rvert}\left(\sum_{i=1}^{\lvert Z\vert} {d_i^{+}}^{2}\right)^{\frac{1}{2}},
\end{equation}
where $Z$ represents the Pareto front or reference set of points. ${d_i}^{+}$ represents the modified distance from $Z$ to the nearest reference point in $X$, which is the solution vector achieved by the algorithm. The IGD+ metric provides the average distance from $Z$ to $X$'s closest points, where the GD+ metric measures the average distance from $X$ to the nearest point in $Z$.

\subsubsection{Meta-heuristic Optimisation Technique Selection}
\begin{table}[htbp]
\caption{PSA System Meta-heuristic Performance Comparison}
\centering
\begin{tabular}{ llr } 
\toprule
    \textbf{Metric} &
    \textbf{Implementation}  & 
    \textbf{Value}   \\
\bottomrule
\toprule
\multirow{4}{*}{\makecell[lt]{HV}}
& Matlab Reference (NSGA-\Romannum{2})  & 0.0067$\pm$0.0005 \\ 
& NSGA-\Romannum{3}                     & 0.0068$\pm$0.0006 \\ 
& C-TAEA                     & 0.0071$\pm$0.0007 \\
& NSGA-\Romannum{2}                     & \textbf{0.0066$\pm$0.0004} \\
\midrule
\multirow{4}{*}{\makecell[lt]{GD+}}
& Matlab Reference (NSGA-\Romannum{2})  & -\\ 
& NSGA-\Romannum{3}                     &   0.0002$\pm$  0.0002 \\ 
& C-TAEA                     &   0.0005$\pm$  0.0005 \\
& NSGA-\Romannum{2}                     &   \textbf{0.0002$\pm$  0.0001} \\ 
\midrule
\multirow{4}{*}{\makecell[lt]{IGD+}}
& Matlab Reference (NSGA-\Romannum{2})& -\\ 
& NSGA-\Romannum{3}                   &   \textbf{0.0040$\pm$ 0.0033} \\ 
& C-TAEA                   &   0.0086$\pm$ 0.0113 \\
& NSGA-\Romannum{2}                   &   0.0059$\pm$ 0.0016 \\  
\bottomrule
\label{tab:psa-opt-comp}
\end{tabular}
\end{table}

A similar approach taken for the CPS systems is implemented for the PSA system to compare various multi-objective optimisation techniques to determine the best performing technique. We include the results for the original Matlab NSGA-\Romannum{2} as a reference and to provide a baseline for the calculation of the metrics GD+ and IGD+. The NSGA-\Romannum{2} and Matlab Reference implementations achieved the same HV values indicating that the NSGA-\Romannum{2} implementation successfully achieved its intent of replicating the Matlab Reference implementation as evidenced in Table~\ref{tab:psa-opt-comp}. The results summarised in Table~\ref{tab:psa-opt-comp} support the use of NSGA-\Romannum{2} by Original Authors Yancy-Caballero \textit{et al.}~\cite{yancy2020process}. Further, these results support our decision to use NSGA-\Romannum{2} as the underlying model for exploring our surrogate assisted methods.

\subsubsection{Surrogate Assisted NSGA-\Romannum{2} Algorithm}\label{algorithm_psa}

The Surrogate Assisted NSGA-\Romannum{2} (SA-NSGA) algorithm that was executed to optimise the PSA system is represented in Figure \ref{algo_psa}. The warm start random population of size 800 is reduced to the 60 best candidates and used as the initial population. The surrogate machine learning models (Random Forest) for the \textit{purity} and \textit{recovery} objectives are both initially trained and parameterised using the random warm start population (an initial random population generated for the intent of training the surrogate model). To handle the PSA system's multi-objective nature, for the surrogate modelling component of the optimisation, two surrogate models for each of the \textit{purity} and \textit{recovery} fitness values were used ~\cite{beck2015multi}.

The initial population is sorted and ranked according to the non-dominated sorting technique using Pareto fronts and the crowding distance metrics (estimate of the surrounding solution density for each gene \cite{raquel2005effective}). The top sixty genes are selected for Tournament Selection. Intermediate crossover and Gaussian mutation are applied with the rate of $\frac{2}{6}$. The offspring are either evaluated using the simulation or the surrogate model based on the divergence criteria. The terminating criterion for the GA is sixty generations. 

The deviation between the surrogate models and the simulation model is managed by evaluating the top 15\% of the ranked population using both the simulation and surrogate models and retraining the surrogate model if the difference is greater than 1$\sigma$. The function evaluation is done in \textsc{Matlab} and the results are sent back to python using the Python \texttt{matlab.engine} library to integrate the two platforms. It is important to note that the integration of these platforms does introduce an additional computational cost.

The differences between the SA-GA algorithm detailed in Section \ref{algorithm_cps} includes the population size, the sorting, selection, crossover and mutation techniques and the number of generations. These adjustments were made to handle the change from a single to a multi-objective system and replicate the approach taken by Yancy-Caballero \textit{et al.}~\cite{yancy2020process}.

\tikzstyle{startstop} = [rectangle, rounded corners, minimum width=2cm, minimum height=1cm,text centered, draw=ashgrey, drop shadow, fill=ashgrey!50,]
\tikzstyle{io} = [trapezium, trapezium left angle=70, trapezium right angle=110, minimum width=2cm, minimum height=1cm, text centered, draw=black, drop shadow]
\tikzstyle{process} = [rectangle, minimum width=2cm, minimum height=1cm, text centered, text width=3cm, draw=orange, drop shadow, fill = orange!20]
\tikzstyle{process2} = [rectangle, minimum width=2cm, minimum height=1cm, text centered, text width=2cm, draw=black, drop shadow, fill = white]
\tikzstyle{decision} = [diamond, minimum width=3cm, minimum height=1cm, text centered, draw=darkred, drop shadow, fill = darkred!20]
\tikzstyle{arrow} = [thick,->,>=stealth]
\tikzset{Tank/.style={draw=black,fill=white,thick,rectangle,rounded corners=10pt,minimum width=0.75cm,minimum height=1.5cm,text width=0.75cm,align=center, drop shadow}}

\tikzstyle{line} = [draw, -stealth]
\begin{figure}[htbp]
\centering
\resizebox{0.43\textwidth}{!}{
\begin{tikzpicture}[node distance=1cm,every node/.style={fill=white}, align=center]
\node (start) [startstop,xshift=3cm] {Start};
\node (pro1) [process, below of =start, yshift=-1cm] {Rank Population (Pareto Rank \& Crowding Distance)};
\node (pro2) [process, below of =pro1, yshift=-1cm] {Select Population from Ranked Population};
\node (pro3) [process, below of =pro2, yshift=-1cm] {Generate Population Tournament Selection};

\node (pro4) [process, below of =pro3, yshift=-1cm] {Generate Offspring Crossover and Mutate};
\node (pro5) [process, below of =pro4, yshift=-1cm] {Evaluate Offspring with Surrogate Model};
\node (pro6) [process, below of =pro5, yshift=-1cm] {Evaluate Elite Population with Simulation};
\node (pro7) [process, below of=pro6,  yshift=-1cm] {Evaluate Surrogate; Check if Outside $1\sigma$};
\node (dec2) [decision, below of=pro7, yshift=-1.75cm] {Divergence\\Criteria};
\node (pro8) [process, left of =dec2,  xshift=-3.7cm] {Retrain Surrogate Model};
\node (inv1) [ right of=dec2, xshift=3.7cm] {};
\draw [arrow] (start) -- (pro1);
\draw [arrow] (pro1) -- (pro2);
\draw [arrow] (pro2) -- (pro3);
\draw [arrow] (pro3) -- (pro4);
\draw [arrow] (pro4) -- (pro5);
\draw [arrow] (pro5) -- (pro6);
\draw [arrow] (pro6) -- (pro7);
\draw [arrow] (pro7) -- (dec2);
\draw [arrow] (dec2) -- node[above]{Yes}(pro8);
\path [line] (dec2.east) node [anchor =south west] {No} --+(1cm,0)  |-(pro1);
\draw[arrow] (pro8.north) -- ++(0,7.35) -- ++(0,6.875) --(pro1.west);
\end{tikzpicture}
}
\caption{Flow Chart for Surrogate Assisted Optimisation Algorithm using NSGA-\Romannum{2}}
\label{algo_psa}
\end{figure}
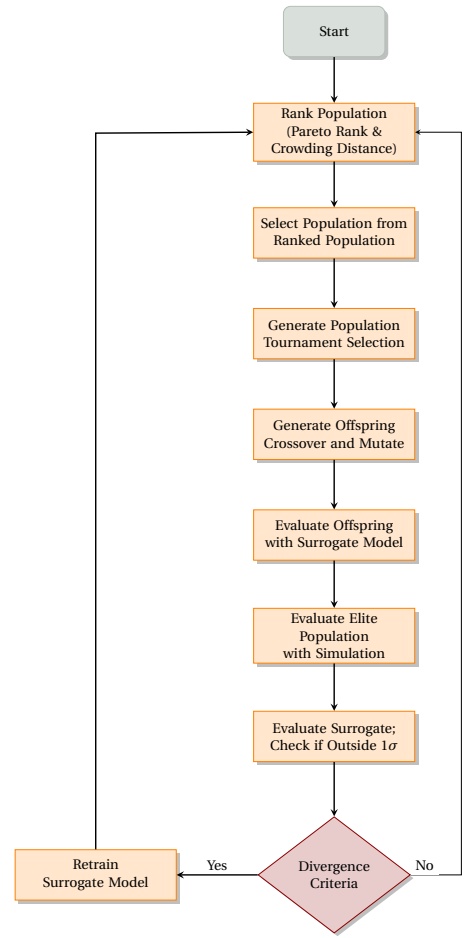

\subsubsection{Selection}

The SA-GA~\cite{art:stander2020,stander2020extended} has been adjusted to substitute the existing sorting technique with the NSGA-\Romannum{2} non-dominated sorting methodology. The sorting algorithm is detailed in Algorithm~\ref{NSGA-Sort}~\cite{deb2002fast}. 

\begin{algorithm}
  \caption{Fast Non-Dominated Sort (P) \cite{deb2002fast}}\label{NSGA-Sort}
  \begin{algorithmic}[1]
        \For{each $p \in P$}
            \State $S_p = \emptyset$
            \State $n_p = 0$
            \For{each $q \in P$} 
                \If{$p<q$}
                    \State $S_p = S_p \cup {q}$
                \ElsIf{$q<p$}
                    \State $n_p = n_p + 1$
                \EndIf
            \EndFor
            \If{$n_p = 0$}
                \State $p_{rank} = 1$
                \State $F_1 = F_1 \cup {q}$
            \EndIf
        \EndFor
        \State $i = 1$
        \While{$F_i\not= \emptyset$}
        \State $Q = \emptyset$
            \For{each $p \in F_i$}
                \For{each $q \in S_p$}
                    \State $n_q = n_q - 1$
                    \If{$n_q = 0$}
                        \State $q_{rank} = i + 1$
                        \State $Q = Q \cup {q}$
                    \EndIf
                \EndFor
                \State $i = i+1$
                \State $F_i = Q$
            \EndFor
        \EndWhile
  \end{algorithmic}
\end{algorithm}

The sorting algorithm executes by iterating over all the solutions in the population. It uses two parameters, namely the domination count ($n_p$), which represents the number of solutions that dominates the current solution, and the solution set that the current solution dominates ($S_p$). The first non-dominated front includes solutions with a domination count of zero. For each solution $p$ with a zero domination count, the algorithm iterates through its solutions $q$ from the set $S_p$ and reduces its domination count by one. When the domination count of any solution $q$ gets to zero, the solution is assigned to a separate list $Q$. These solutions represent the second non-dominated front \cite{deb2002fast}.

Once the non-dominated sorting is applied, the sorting algorithm's non-domination ranking output is used in combination with each solution's crowding distance metric to sort the population. The next phase of the SA-NSGA is the tournament selection process which is executed to generate a population for the offspring generation process. The tournament selection process involves the random sampling of a set number of individuals, with replacement from the existing population with the intent of comparing their fitness and selecting the best individual for the new population \cite{fang2010review}. The number of individuals sampled at every iteration is two, and the population size is sixty, which results in 120 tournaments taking place to achieve a population size of sixty.

\subsubsection{Crossover and Mutation} \label{crossover_mutation_psa}

The population generated by the tournament selection process is then used for crossover using the intermediate crossover technique. This technique creates offspring by taking a weighted average of the parents using a ratio value to specify the weights. The offspring are a function of two parents, parent1 and parent2. The first child is given by:
\begin{equation}
\label{eq:crossover_psa_child1}
\text{Child 1} = \text{Parent 1} + \text{Rand} \times \text{Ratio} \times \text{(Parent 2 - Parent 1)}
\end{equation}
and the second by:
\begin{equation}
\label{eq:crossover_psa_child2}
\text{Child 2} = \text{Parent 2} - \text{Rand} \times \text{Ratio} \times \text{(Parent 2 - Parent 1)}
\end{equation}
where the Ratio is a (uniform) randomly generated weight used to determine the change in genetic structure of the child from the parent's. The probability of a child changing its genetic structure from its parent is based on the Rand vector, a set of binary values, the same length as the number of decision variables. The Rand vector is determined by initially having a generated vector set of random numbers and checking if they are less than the specified fraction (crossover rate) of $\frac{2}{6}$.

\begin{figure}[htbp]
\centering
\begin{tikzpicture}[node distance=1cm,align=center, scale = 1.25]
\draw[drop shadow, fill=white] (0,1) rectangle (0.5,1.5) node [xshift = -9, yshift = -8] {\footnotesize 5};
\draw[drop shadow, fill=white] (0.5,1) rectangle (1,1.5) node [xshift = -9, yshift = -8] {\footnotesize 20};
\draw[fill=gray!30,drop shadow] (1,1) rectangle (1.5,1.5) node [xshift = -9, yshift = -8] {\footnotesize 0.2};
\draw[fill=white,drop shadow] (1.5,1) rectangle (2,1.5) node [xshift = -9, yshift = -8] {\footnotesize 1.8};
\draw[fill=gray!30,drop shadow] (2,1) rectangle (2.5,1.5) node [xshift = -9, yshift = -8] {\footnotesize 0.9};
\draw[fill=gray!30,drop shadow] (2.5,1) rectangle (3,1.5) node [xshift = -9, yshift = -8] {\footnotesize 0.15};
\node [left] at (0,1.25 ) {Parent 2};
\draw[fill=white,drop shadow] (0,2) rectangle (0.5,2.5) node [xshift = -9, yshift = -8] {\footnotesize 9};
\draw[fill=white,drop shadow] (0.5,2) rectangle (1,2.5) node [xshift = -9, yshift = -8] {\footnotesize 100};
\draw[fill=gray!30, drop shadow] (1,2) rectangle (1.5,2.5) node [xshift = -9, yshift = -8] {\footnotesize 0.5};
\draw[fill=white, drop shadow] (1.5,2) rectangle (2,2.5) node [xshift = -9, yshift = -8] {\footnotesize 1.5};
\draw[fill=gray!30, drop shadow] (2,2) rectangle (2.5,2.5) node [xshift = -9, yshift = -8] {\footnotesize 0.8};
\draw[fill=gray!30, drop shadow] (2.5,2) rectangle (3,2.5) node [xshift = -9, yshift = -8] {\footnotesize 0.25};
\node [left] at (0,2.25 ) {Parent 1};
\node[fill=ashgrey!50, style={single arrow, draw=none, rotate=270}] at (1.5, 0.125) {crossover};
\draw[fill=white,drop shadow] (0,-1.25) rectangle (0.5,-.75) node [xshift = -9, yshift = -8] {\footnotesize 9};
\draw[fill=white,drop shadow] (0.5,-1.25) rectangle (1,-.75) node [xshift = -9., yshift = -8] {\footnotesize 100};
\draw[fill=white,drop shadow] (1,-1.25) rectangle (1.5,-.75) node [xshift = -9, yshift = -8] {\footnotesize 0.29};
\draw[fill=white,drop shadow] (1.5,-1.25) rectangle (2,-.75) node [xshift = -9, yshift = -8] {\footnotesize 1.5};
\draw[fill=white,drop shadow] (2,-1.25) rectangle (2.5,-.75) node [xshift = -9, yshift = -8] {\footnotesize 0.59};
\draw[fill=white,drop shadow] (2.5,-1.25) rectangle (3,-.75) node [xshift = -9, yshift = -8] {\footnotesize 0.45};
\node [left] at (0,-1. ) {Child 1};
\node [left] at (3, 0.125) {Equation 4 \\ Crossover \\ Example};


\end{tikzpicture}
\caption{An Example of the Intermediate Crossover Operation for two potential parent candidates and a resulting new chromosome}
\label{fig:crossover_psa}
\end{figure}
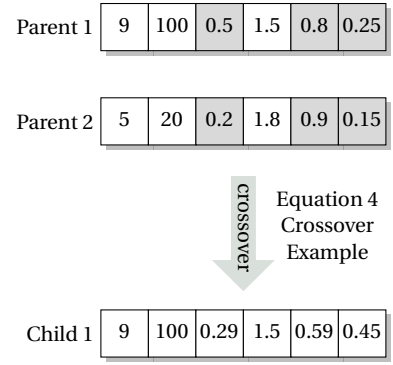

Figure \ref{fig:crossover_psa} is an example of the crossover procedure for PSA optimisation. The grey blocks represent the parent elements that will be changed, and the white blocks represent the elements that will remain from chromosome 1 in child 1.

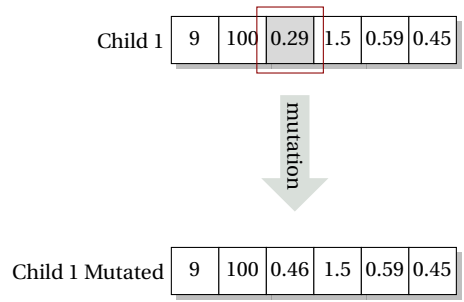
\begin{figure}[htbp]
\centering
\begin{tikzpicture}[node distance=1cm,align=center, scale = 1.25]
\draw[fill=white,drop shadow] (0,1) rectangle (0.5,1.5) node [xshift = -9, yshift = -8] {\footnotesize 9};
\draw[fill=white,drop shadow] (0.5,1) rectangle (1,1.5) node [xshift = -9, yshift = -8] {\footnotesize 100};
\draw[fill=gray!30, drop shadow] (1,1) rectangle (1.5,1.5) node [xshift = -9, yshift = -8] {\footnotesize 0.29};
\draw[fill=white,drop shadow] (1.5,1) rectangle (2,1.5) node [xshift = -9, yshift = -8] {\footnotesize 1.5};
\draw[fill=white,drop shadow] (2,1) rectangle (2.5,1.5) node [xshift = -9, yshift = -8] {\footnotesize 0.59};
\draw[fill=white,drop shadow] (2.5,1) rectangle (3,1.5) node [xshift = -9, yshift = -8] {\footnotesize 0.45};
\node [left] at (0,1.25 ) {Child 1};
\node[fill=ashgrey!50, style={single arrow, draw=none, rotate=270}] at (1.3, 0.125) {mutation};
\draw[fill=white,drop shadow] (0,-1.45) rectangle (0.5,-.95) node [xshift = -9, yshift = -8] {\footnotesize 9};
\draw[fill=white,drop shadow] (0.5,-1.45) rectangle (1,-.95) node [xshift = -9., yshift = -8] {\footnotesize 100};
\draw[fill=white,drop shadow] (1,-1.45) rectangle (1.5,-.95) node [xshift = -9, yshift = -8] {\footnotesize 0.46};
\draw[fill=white,drop shadow] (1.5,-1.45) rectangle (2,-.95) node [xshift = -9, yshift = -8] {\footnotesize 1.5};
\draw[fill=white,drop shadow] (2,-1.45) rectangle (2.5,-.95) node [xshift = -9, yshift = -8] {\footnotesize 0.59};
\draw[fill=white,drop shadow] (2.5,-1.45) rectangle (3,-.95) node [xshift = -9, yshift = -8] {\footnotesize 0.45};
\node [left] at (0,-1.2 ) {Child 1 Mutated};

\draw [darkred] (0.9, 1.6) rectangle (1.6, 0.9);

\end{tikzpicture}
\caption{An Example of the Gaussian Mutation Operation for the potential candidate used by the GA}
\label{fig:mutation_psa}
\end{figure}

The Gaussian mutation technique adds a random number from the Gaussian distribution, with a mean of 0 and standard deviation set by two parameters: the shrink and scale parameters, to each child's entry. The mutation is applied at a rate of $\frac{2}{6}$. An example of the mutation procedure is shown in Figure \ref{fig:mutation_psa}.

At the end of both of these procedures, the decision variables are checked against their bounds. If the decision variables lie above or below their respective bounds, they are substituted with either the maximum or the minimum values, respectively.

\subsubsection{Genetic Structure}\label{motivating_genetic_structure}

The decision variables detailed in Section \ref{sec:PSA} formed part of the chromosomes that made up the GA population. The genetic structure of the chromosomes for the PSA system optimisation are summarised in Table \ref{PSAGene}. The genes were encoded with real numbers within the bounds specified in Table \ref{PSAGene}.

\begin{table}[htbp]
\caption{Genetic Structure for Gene 1 to 6}
\centering
\resizebox{.975\columnwidth}{!}{
\begin{tabular}{ rllc } 
\toprule
{\textbf{\#}} & {\textbf{Description}} & {\textbf{Units}}& {\textbf{Bounds}}\\
\bottomrule
\toprule
1 & Adsorption Pressure  & Bar & [1,10]\\ 
2 & Time of Adsorption & Seconds & [10, 1000] \\  
3 & Light Product Reflux Ratio & - & [0.01, 0.99]\\
4 & Feed Velocity & m/s & [0.1, 2]\\
5 & Heavy Product Reflux Ratio & - & [0,1]\\ 
6 & Purge Pressure & Bar & [0.1, 0.5]\\
\bottomrule
\end{tabular}
}
\label{PSAGene}
\end{table}

\subsubsection{Surrogate Model} \label{surrogate_psa}

\begin{table}[htbp]
\caption{Machine Learning Surrogate Comparison - PSA System Performance Comparison}
\centering
\begin{tabular}{ llr } 
\toprule
    \textbf{Metric} &
    \textbf{ML Technique}  & 
    \textbf{Value}   \\
\bottomrule
\toprule
\multirow{2}{*}{\makecell[lt]{HV}}
& Random Forest (Surrogate)     & 1.303\\
& Decision Tree (Surrogate)     & \textbf{0.681}\\
\midrule
\multirow{2}{*}{\makecell[lt]{GD+}}
& Random Forest (Surrogate)     & \textbf{0.500}\\
& Decision Tree (Surrogate)     & 6.000\\
\midrule
\multirow{2}{*}{\makecell[lt]{IGD+}}
& Random Forest (Surrogate)     &  \textbf{1.357}\\
& Decision Tree (Surrogate)     &  4.021\\
\bottomrule
\label{tab:ml-perf-comp-psa}
\end{tabular}
\end{table}

A similar approach implemented for the CPS systems was made for the PSA system to test machine learning techniques as surrogate models to determine the best performing algorithm. We select the two best performing methods for CPS, Decision Tree and Random Forest. The Random Forest and Decision Tree surrogate assisted implementation are included in Table~\ref{tab:ml-perf-comp-psa}. These metrics represent the relative average performance increase/decrease achieved across all twenty experiments for the final generation (60). We use a Non-surrogate NSGA-\Romannum{2} (Direct) as the reference.  It is important to note that for all the metrics summarised in this table, the best values have been highlighted in bold. The Random Forest surrogate assisted implementation has a worse HV than the Decision Tree Surrogate but was able to out-perform for both GD+ and IGD+. Based on these results the Random Forest algorithm detailed in Section \ref{surrogate_cps} was selected for the PSA optimisation for both \textit{purity} and \textit{recovery} and was selected for further analysis. The details of the hyper-parameters can be found in Section~\ref{surrogate_cps}.

\subsection{Software and Hardware Details}

The optimisation extension component of the study made use of the \textit{pymoo} library, which includes a large range of single and multi-objective optimisation techniques for python~\cite{pymoo}. The simulation models and the SA-GA Algorithm for the optimisation of the motivating examples were all implemented in Python. Three different implementations were executed for the PSA optimisation case. The first implementation is the direct implementation which was executed in \textsc{Matlab} using the source code provided by Yancy-Cabellero \textit{et al.}~\cite{yancy2020process}. The second is the reference implementation involving the replication of the algorithm used in the direct implementation in python. The final is the surrogate assisted implementation, which introduced surrogate models into the reference approach as substitutes for the simulation model. The surrogate assisted implementation is focused on the potential computational efficiency improvements that can be achieved by using surrogate models. The performance metrics detailed in Section \ref{psa-perf} were implemented in Python using \textit{pymoo}. \textit{Pymoo} is the multi-objective optimisation library used to calculate the HV, GD+ and the IGD+. The HV metric within the \textit{pymoo} library was implemented using the algorithm developed by Distributed Evolutionary Algorithms in Python (DEAP). All the implementations for the optimisation of both the primary and motivating examples are available on a Github~\cite{Liezl2021}.

\section{Results}
\label{sec:results}

\subsection{Motivating Examples - CPS Systems}

Thirty repetitions of each experiment were completed for each system, using two modes of optimisation. The first mode involved only the simulation model as the evaluation platform. The second mode involved the integration of both the simulation and the surrogate model. For the simulation only mode, the primary metric used to determine performance was the mean revenue for the population for both CPS-1 and CPS-2. Two primary metrics were used to evaluate the performance of the surrogate assisted mode for CPS-1. The first metric was the mean revenue for the population evaluated by the surrogate and simulation models. The second metric was the mean revenue for the elite population evaluated by both the simulation and surrogate model. CPS-2's performance for the surrogate assisted mode was illustrated by comparing the system's deterministic and stochastic version. The deterministic results included the mean of the max revenues, and the stochastic results included the mean revenues at different values for the maintenance hours decision variable. The maintenance hours were fixed at their maximum and minimum values. The revenues of the two experiments were compared. The first experiment had fixed the maintenance hours and the second experiment had the GA select the optimal maintenance hours. 

The remainder of this section has been subdivided into the results for CPS-1 and CPS-2.

\subsubsection{CPS-1 Results}\label{sec:problem1Results}

The results from the simulation only mode for CPS-1 illustrated that the GA optimised across a more complex stochastic system with increased decision variables.

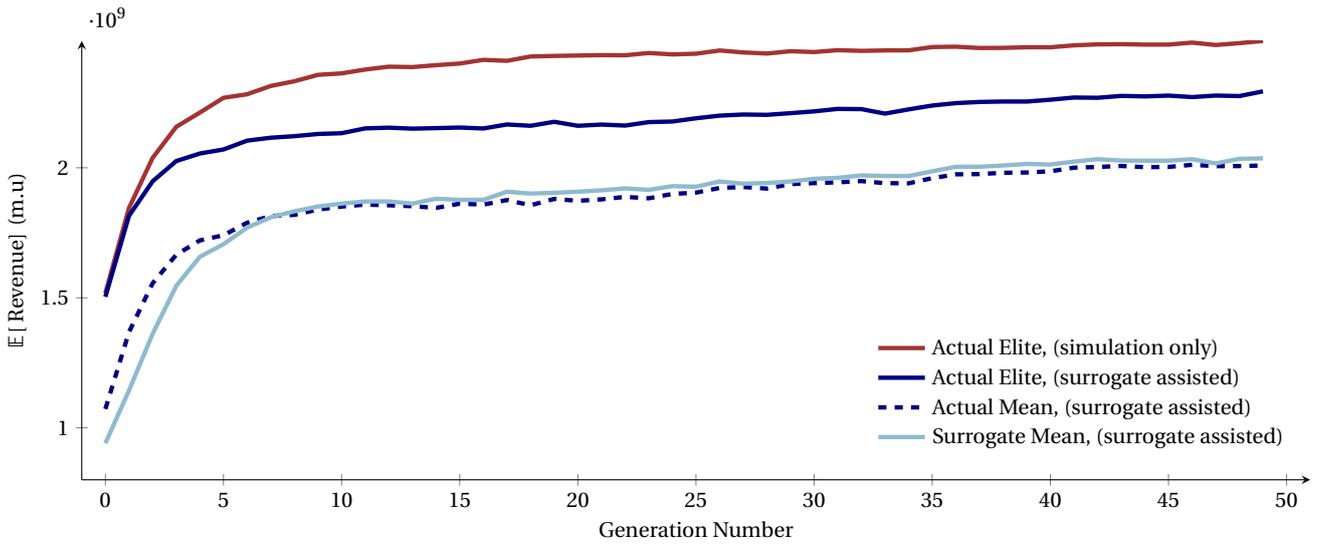
\begin{figure*}[htbp]
\centering
\resizebox{\textwidth}{!}{
\begin{tikzpicture}[scale=.75]
\begin{axis}[
    xmin = -1,  xmax = 51,  
    ymin = 0.8e9, 
    ylabel=$\mathbb{E\left[\text{ Revenue}\right]\text{ (m.u)}}$ ,
    xlabel=Generation Number,
    axis lines = left,
    legend cell align={left},
    legend style={at={(0.99,0.05)},
		anchor=south east,legend columns=1, draw=none},
    width=\textwidth,
    height=\axisdefaultheight
    ]

\addplot [darkred!80, ultra thick]
     table [x = Generations, y = Mean_Revenue_Elite_Mean, col sep=comma ] {Parallel_System_simulation_30_exp_Processed.txt};
\addlegendentry{Actual Elite, (simulation only)}

\addplot [navyblue, ultra thick]
   table [ x = Generations, y = Mean_Revenue_Elite_Mean, col sep=comma ] {Parallel_System_surr_30_exp_Processed.txt};
\addlegendentry{Actual Elite, (surrogate assisted)}

\addplot [navyblue, dashed, ultra thick]
    table [x = Generations, y = Mean_Revenue_Mean, col sep=comma ] {Parallel_System_surr_30_exp_Processed.txt};
\addlegendentry{Actual Mean, (surrogate assisted)}

\addplot [moonstoneblue!80,  ultra thick]
    table [x = Generations, y = Mean_Revenue_Surrogate_Mean, col sep=comma ] {Parallel_System_surr_30_exp_Processed.txt};
\addlegendentry{Surrogate Mean, (surrogate assisted)}

\end{axis}
\end{tikzpicture}
}
\caption{Actual mean revenue, surrogate mean revenue,  actual elite mean revenue for the surrogate assisted mode and the actual elite mean revenue for the simulation only mode (CPS-1)}
\label{figprob1surrmeanactmean}
\end{figure*}

Figure \ref{figprob1surrmeanactmean} illustrates the SA-GA algorithm results for the optimisation of the CPS-1 system. The actual lines (solid dark red, solid dark blue and dotted dark blue) represent the results where the simulation model was used to evaluate the population. The surrogate line (solid light blue) represents where the surrogate model was used to evaluate the population. The solid dark red and blue lines represent the mean actual revenues for the elite population in the simulation only and surrogate assisted optimisation modes, respectively. The solid light blue and dotted dark blue lines represent the mean actual and surrogate revenues for the entire population in the surrogate assisted optimisation modes. The performance of the SA-GA algorithm is evident from Figure \ref{figprob1surrmeanactmean}. Not only is the GA able to optimise across a more complex stochastic system within the simulation only mode, but it is also improving the elite population across generations for both the simulation only and surrogate assisted implementations. The surrogate assisted algorithm can learn the more complex CPS in CPS-1 with an accuracy of 90\% and yields a speedup of 1.7 times over the simulation only mode. The CPU times of these implementations are summarised in Table \ref{tab:motivating-cpu}

\subsubsection{CPS-2 Results}\label{sec:FeedbackResults}

The optimisation of CPS-2 in the simulation only implementation yielded positive results indicating the GA's robustness to finding optimal solutions in an increasingly dynamic and complex system that produces a varying result for the same set of input variables due to the stochasticity introduced into the system.

Figure \ref{figprob2fixedsystem} includes the results for three different experiments completed on the deterministic version of CPS-2. The dark and light red lines represent the revenues when the maintenance hours are fixed at the maximum and minimum bound of 1314 and 0 hours. The dotted red lines represent the revenue of the experiments where the GA could select and optimise the maintenance hours. These three different experiments have also been completed for the stochastic version of CPS-2~\cite{stander2020extended}.

\pgfplotsset{ every non boxed x axis/.append style={x axis line style=-}}
\begin{figure*}[htbp]
\centering
\resizebox{\textwidth}{!}{
\begin{tikzpicture}[scale=.75]
\begin{axis}[
    xmin  = -1,  xmax = 51,  
    ymin  =  0.85e8, 
    ylabel=$\mathbb{E\left[\text{Max Revenue}\right]}\text{ (m.u)}$,
    xlabel=Generation Number,
    axis lines = left,
    width=\textwidth,
    height=\axisdefaultheight
    ]
    
\addplot [darkred, ultra thick]
table [ x = Generations, y = Max_Revenue_Surrogate_Mean, col sep=comma ] {Feedback_Loop_Fixed_Max.txt} ;\label{plot1}

\addplot [darkred!50, ultra thick]
table [ x = Generations, y = Max_Revenue_Surrogate_Mean, col sep=comma ] {Feedback_Loop_Fixed_Min.txt} ;\label{plot2}

\addplot [darkred!80,  dashed, ultra thick]
table [ x = Generations, y = Max_Revenue_Surrogate_Mean, col sep=comma ] {Feedback_Loop_Surr_30_fixed_Processed.txt} ;\label{plot3}

\end{axis}

\begin{axis}[
    axis y line=right,
    axis x line=none,
    ymax = 1200,
    ymin = 500,
    xmax = 51,
    axis y line shift=0.2,
    ylabel= Maintenance Hours (h),
    legend cell align={left},
    legend style={at={(0.99,0.05)},
		anchor=south east,legend columns=1, draw=none},
    width=\textwidth,
    height=\axisdefaultheight
]
\addlegendimage{/pgfplots/refstyle=plot1}\addlegendentry{(m.u) @ Maximum Maintenance Hours}
\addlegendimage{/pgfplots/refstyle=plot2}\addlegendentry{(m.u) @ Minimum Maintenance Hours}
\addlegendimage{/pgfplots/refstyle=plot3}\addlegendentry{(m.u) @ Variable Maintenance Hours}

\addplot [darkgoldenrod, ultra thick]
table [ x = Generations, y = mean_hours_maintained, col sep=comma] {Population_Variables_30_reps_Fixed_System_processed_2.txt} ;\label{plot4}

\addlegendentry{Maintenance Decision Variable (h)}
\end{axis}
\end{tikzpicture}
}
\caption{Mean of the max revenue of the deterministic system with constant and variable maintenance hours (CPS-2)}
\label{figprob2fixedsystem}
\end{figure*}
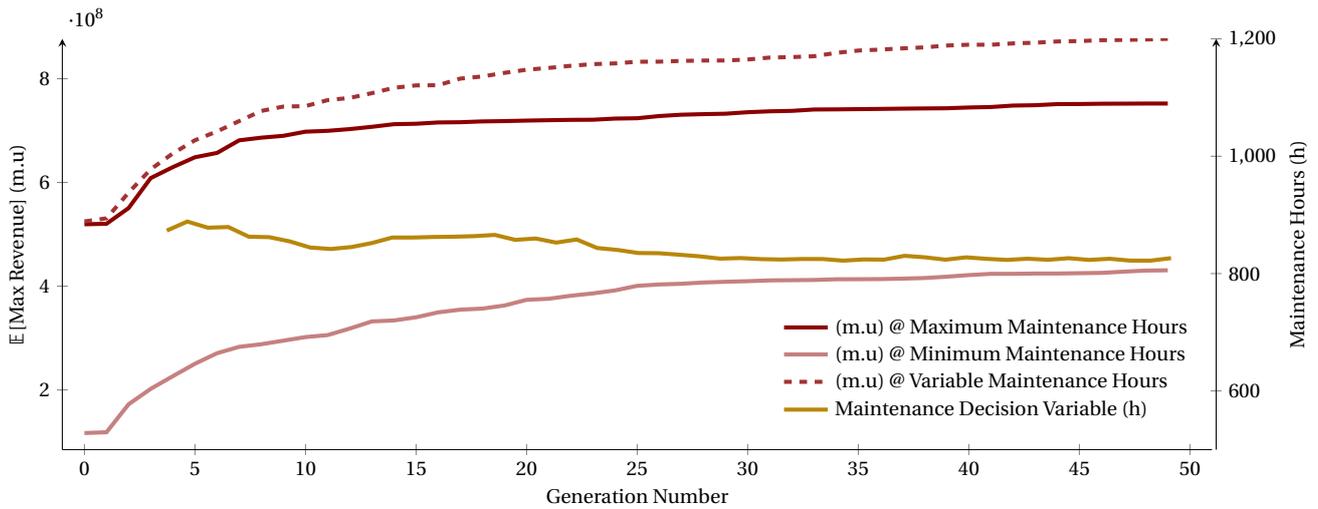

These results indicate that the surrogate assisted implementation improved the revenue of CPS-2 despite the additional complexity added. This implementation also yielded a speedup of 1.84 times the simulation only mode, and the surrogate model was able to predict the revenue from CPS-2 at an overall accuracy of 69\%. The GA was able to handle the additional complexity added by the system's feedback component, indicating its robustness towards this type of component.The detailed CPU times are summarised in Table \ref{tab:motivating-cpu}.

\subsubsection{CPS System CPU Times}\label{sec:cpu-times-cps}

The details of the CPU times for the GA (Direct) and Random Forest SA-GA are summarised in Table \ref{tab:motivating-cpu}. These results illustrate the significant speedup achieved by the Random Forest surrogate model.

\begin{table}[htbp]
\caption{CPS CPU Times}
\centering
\begin{tabular}{ llr } 
\toprule
    \textbf{System} &
    \textbf{ML Technique}  & 
    \textbf{\makecell[r]{CPU Time\\(seconds)}}   \\
\bottomrule
\toprule
\multirow{2}{*}{\makecell[l]{CPS 1}}
& GA (Direct)               & $4.0 \times 10^{4}$        \\  
& Random Forest (Surrogate) & $\mathbf{2.4 \times 10^{4}}$  \\  
\midrule
\multirow{2}{*}{\makecell[l]{CPS 2\\(Stochastic)}}
& GA (Direct)               & $1.4 \times 10^{4}$     \\  
& Random Forest (Surrogate) & $\mathbf{7.6 \times 10^{3}}$ \\  
\bottomrule
\label{tab:motivating-cpu}
\end{tabular}
\end{table}

\subsection{Primary Example - PSA System}

A set of twenty experiments were run for the Matlab Reference and Random Forest surrogate assisted implementations. The NSGA-\Romannum{2} (Direct) implementation was executed for five runs due to this implementation focusing on replicating the Matlab Reference implementation and having significantly higher computational cost.  

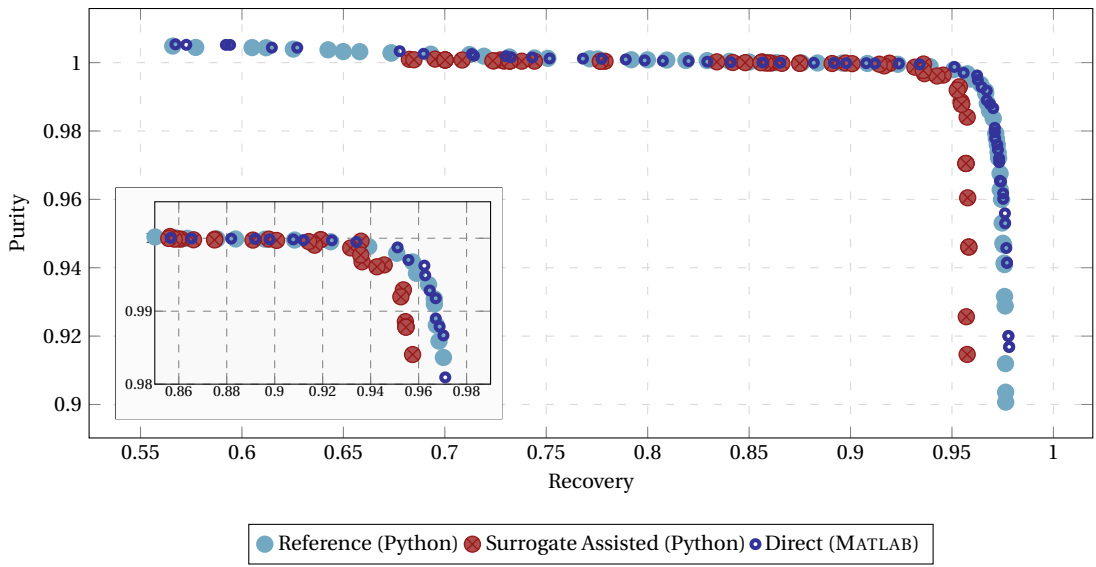
\begin{figure*}[htbp]
\centering
\begin{tikzpicture}
\begin{pgfonlayer}{background}
\begin{axis}[
    ylabel=Purity,
    xlabel=Recovery,
    xmajorgrids=true,
    ymajorgrids=true,
    grid style={gray!30, loosely dashed},
    width=.85\textwidth,
    height=\axisdefaultheight,
    legend style={at={(0.5,-0.2)},
    	anchor=north,legend columns=3}
		]
\addplot [only marks,mark size = 3, mark options={
                    draw= moonstoneblue,
                    fill = moonstoneblue!100,
                    line width = 0.5pt
                }]
    table [x = Purity, y = Recovery, col sep=comma] {Direct_Opt_PSA_Python.txt};
\addlegendentry{Reference (Python)}

\addplot [only marks, mark=otimes*,mark size = 3, mark options={
                    draw= darkred!90,
                    fill = darkred!70,
                    line width = 0.5pt
                }]
    table [x = Purity, y = Recovery, col sep=comma] {Surr_assisted_PSA.txt};
\addlegendentry{Surrogate Assisted (Python)}

\addplot [only marks,mark=o, mark size = 1.5,mark options={
                    draw = navyblue!80,
                    line width = 1.5pt
                }]
    table [x = Purity, y = Recovery, col sep=comma] {Direct_Opt_PSA_Matlab.txt};
\addlegendentry{Direct (\textsc{Matlab})}
\coordinate (inset) at (axis description cs:0.40,0.55);
\end{axis}
\end{pgfonlayer}
\begin{pgfonlayer}{foreground}
\begin{axis}[
    name=ax2,
    scaled ticks=true,
    at={(inset)},
    anchor= north east,
    small,
    xmajorgrids=true,
    ymajorgrids=true,
    grid style={help lines,dashed},
    width=6cm,
    height=4cm,
    xmin=0.85, xmax=0.99, 
    ymin=0.98, ymax=1.005,
    xtick distance=0.02,
    ytick distance=0.01,
    yticklabel style = {font=\tiny,xshift=0.5ex},
    xticklabel style = {font=\tiny,yshift=0.5ex},
    axis background/.style={fill=gray!5}
    ]
\addplot [only marks,mark size = 3, mark options={
                    draw= moonstoneblue,
                    fill = moonstoneblue!100,
                    line width = 0.5pt
                }]
    table [x = Purity, y = Recovery, col sep=comma] {Direct_Opt_PSA_Python.txt};

\addplot [only marks, mark=otimes*,mark size = 3, mark options={
                    draw= darkred!90,
                    fill = darkred!70,
                    line width = 0.5pt
                }]
    table [x = Purity, y = Recovery, col sep=comma] {Surr_assisted_PSA.txt};

\addplot [only marks,mark=o, mark size = 1.5,mark options={
                    draw = navyblue!80,
                    line width = 1.5pt
                }]
    table [x = Purity, y = Recovery, col sep=comma] {Direct_Opt_PSA_Matlab.txt};
\end{axis}
\end{pgfonlayer}
\begin{pgfonlayer}{main}
\draw[black] ([shift={(-1pt,-6pt)}] ax2.outer south west)
    rectangle    ([shift={(+4pt,+5pt)}] ax2.outer north east);
\fill [gray!3] ([shift={(-1pt,-6pt)}] ax2.outer south west)
    rectangle    ([shift={(+4pt,+5pt)}] ax2.outer north east);
\end{pgfonlayer}

\end{tikzpicture}
\caption{Frontier Comparison between the direct, surrogate assisted and reference implementations (PSA)}
\label{psa_pareto}
\end{figure*}

Three primary quantitative metrics, HV, GD+ and IGD+, were used to illustrate the performance of the various implementations. Each metrics was calculated per generation for all experiments.  For the twenty experiments, the average of the final generation's population generated from the direct implementation was used as the reference vector. The qualitative metric that was used across all the experiment sets was the Pareto frontier for the last generation. This metric was intended to provide visual comparisons of the performances across the different implementations. The final generation frontiers for all three implementations are represented in Figure~\ref{psa_pareto}.

The NSGA-\Romannum{2} (Direct) implementation has achieved a similar shape and range to the Matlab Reference implementation from visual inspection of these frontiers. The frontier of the Random Forest surrogate assisted implementation also follows the same shape as the Matlab Reference implementation. To further validate the results illustrated by the Pareto frontiers, quantitative measures have been used. 

\begin{figure*}[htbp]
\centering
\begin{tikzpicture}
\begin{pgfonlayer}{background}
\begin{axis}[
    xmin = -1,  xmax = 61,  
    ymin = 0.0, ymax = 0.2,  
    ylabel= Minimum Hypervolume,
    xlabel=Generation Number,
    axis lines = left,
    legend cell align={left},
    legend style={at={(0.95,0.95)},
		anchor=north east,legend columns=1, draw=none},
    width=\textwidth,
    height=\axisdefaultheight
    ]
\addplot [moonstoneblue, dashed, thick]
    table [x = generation, y = average_HV_Python, col sep=comma ] {HV_Avg.txt};
\addlegendentry{NSGA-\Romannum{2} (Direct)}

\addplot [navyblue!80, ultra thick]
    table [x = generation, y = average_HV_Matlab, col sep=comma ] {HV_Avg.txt};
\addlegendentry{Matlab Reference}]

\addplot [darkred!90, ultra thick]
     table [x = generation, y = average_HV_Surr, col sep=comma ] {HV_Avg.txt};
\addlegendentry{Random Forest Surrogate Assisted}


\coordinate (inset) at (axis description cs:0.95,0.6);
\end{axis}
\end{pgfonlayer}
\begin{pgfonlayer}{foreground}
\begin{axis}[
    name=ax2,
    scaled ticks=true,
    at={(inset)},
    anchor= north east,
    small,
    xmajorgrids=true,
    ymajorgrids=true,
    grid style={help lines,dashed},
    width=6cm,
    height=4cm,
    xmin=5, xmax=30, 
    ymin=0.004, ymax=0.023,
    yticklabel style = {font=\tiny,xshift=0.5ex},
    xticklabel style = {font=\tiny,yshift=0.5ex},
    axis background/.style={fill=gray!5}
    ]
    
\addplot [moonstoneblue, dashed, thick]
    table [x = generation, y = average_HV_Python, col sep=comma ] {HV_Avg.txt};

\addplot [navyblue!80, ultra thick]
    table [x = generation, y = average_HV_Matlab, col sep=comma ] {HV_Avg.txt};

\addplot [darkred!90, ultra thick]
     table [x = generation, y = average_HV_Surr, col sep=comma ] {HV_Avg.txt};
\end{axis}
\end{pgfonlayer}
\begin{pgfonlayer}{main}
\draw[black] ([shift={(-1pt,-6pt)}] ax2.outer south west)
    rectangle    ([shift={(+4pt,+5pt)}] ax2.outer north east);
\fill [gray!3] ([shift={(-1pt,-6pt)}] ax2.outer south west)
    rectangle    ([shift={(+4pt,+5pt)}] ax2.outer north east);
\end{pgfonlayer}
\end{tikzpicture}
\caption{Mean Hypervolume progression at each generation across repeated experiments (PSA)}
\label{AvgHV}
\end{figure*}
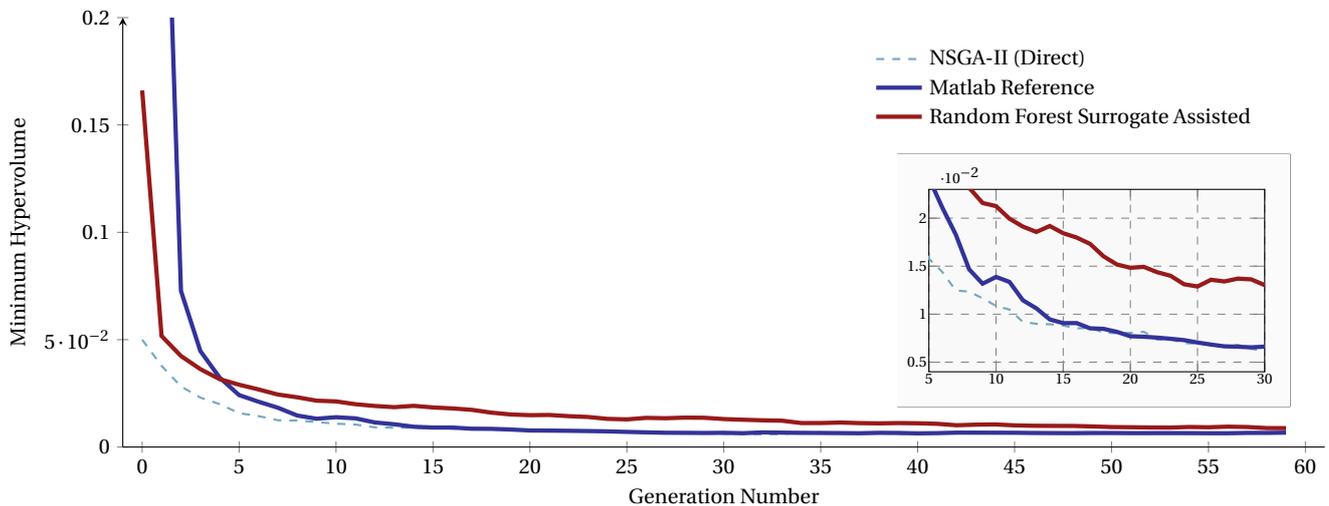

To further investigate each implementation's convergence, the average of the HV per generation is illustrated in Figure \ref{AvgHV} for all implementations. We note no material difference between progressions other than additional noise in the surrogate's solution space in early generations.

Each implementation has a specific convergence rate, and to illustrate this in a comparative nature, the average number of evaluations (generations) to success (AES) for 99.5\%, 99\%, 98.5\% and 98\% HV success rates has been summarised in Table \ref{tab:aes}. The surrogate assisted implementation yields the slowest convergence rate across all the implementations. It is, however, able to achieve the 99.5\% Success Rate of 99.5\% within the maximum number of generations (60). The Surrogate Assisted model requires significantly more generations to achieve the same success rate as the other approaches. However, these additional generations are evaluated using the surrogate model and as such still result in a significant speedup.

\begin{table}[htbp]
\caption{Average Evaluations to Success (AES) for various Success Rates (Number of Generations) for NSGA-\Romannum{2}}
\label{PSAExperiments}
\centering
\begin{tabular}{lrrrr} 
\toprule
    & \multicolumn{4}{c}{\textbf{Success Rate}} \\
    \textbf{Implementation}  & 
    \textbf{98.0\%} & 
    \textbf{98.5\%}&
    \textbf{99.0\%}&
    \textbf{99.5\%}\\
\bottomrule
\toprule
Matlab Reference          & \textit{6} & \textit{7}  & \textit{10} & \textit{16} \\ 
NSGA-\Romannum{2} (Direct)            & \textbf{3} & \textbf{5}  & \textbf{6}  & \textbf{12} \\ 
Random Forest (Surrogate) & \textit{7} & \textit{12} & \textit{22} & \textit{49} \\ 
\bottomrule
\end{tabular}
\label{tab:aes}
\end{table}

From the results above, we have evidence demonstrating that the NSGA-\Romannum{2} (Direct) implementation achieved equivalent results to the Matlab Reference implementation. The Random Forest surrogate assisted technique achieved slightly inferior results to the NSGA-\Romannum{2} (Direct) and Matlab Reference optimisation cases but has significantly reduced the computation cost of the optimisation. The surrogate assisted implementation achieved a 2.7 times speedup compared to the NSGA-\Romannum{2} (Direct) implementation with an accuracy of 91.8\% for the \textit{purity} and 99.1\% for the \textit{recovery}. The details of the CPU times for the Matlab Reference, NSGA-\Romannum{2} (Direct) and the Random Forest surrogate assisted implementations are summarised in Table \ref{tab:psa-cpu}. The Matlab Reference implementation has a shorter CPU time than the NSGA-\Romannum{2} (Direct) due to the computational cost introduced with Python interfacing with Matlab.

\begin{table}[htbp]
\caption{PSA CPU Times}
\centering
\begin{tabular}{ llr } 
\toprule
    \textbf{System} &
    \textbf{ML Technique}  & 
    \textbf{\makecell[r]{CPU Time\\(seconds)}}   \\
\bottomrule
\toprule
\multirow{3}{*}{\makecell[l]{PSA}}
& Matlab Reference                      & $3.9 \times 10^{4}$ \\
& NSGA-\Romannum{2} (Direct)            & $3.5 \times 10^{5}$ \\  
& Random Forest (Surrogate)             & $\mathbf{1.3 \times 10^{5}}$\\  
\bottomrule
\label{tab:psa-cpu}
\end{tabular}
\end{table}

\section{Discussion}

The study illustrated that the best performing meta-heuristic optimisation techniques were the GA for the two motivating examples and the NSGA-\Romannum{2} for the multi-objective PSA system. The small differences in results illustrated that the other algorithms could optimise across these complex systems with marginal differences in performance. The surrogate model comparison results highlighted the strength of the Random Forest algorithm as the best performing machine learning model. The computational time-saving achieved by combining the GA and NSGA-\Romannum{2} meta-heuristic algorithms with the Random Forest surrogate model ranged between a factor of 1.7 - 2.7 times speedup. These speedups were achieved with surrogate model accuracies of 69\% - 99.1\%. The significant computational time reduction as detailed by authors in various research papers \cite{ali2018surrogate,shi2016evolutionary,beck2015multi,anna2017machine,ibrahim2018optimization} was replicated in this study. The NSGA-\Romannum{2} algorithm proved to be the best performing algorithm for the PSA system as supported by previous research \cite{pai2020experimentally,subraveti2019machine,yancy2020process}.

The findings in this study are extremely valuable for Industry. Chemical plant systems are usually stochastic with numerous complexities. The results achieved illustrated that meta-heuristic optimisation algorithms can achieve optimal solutions despite these complexities. The primary benefit of the meta-heuristic optimisation technique is that it does not require problem-specific information and is then easily combined with the surrogate models for computational time savings. The combination of optimising across computationally expensive and complex chemical plant models and the surrogate model substitute for these models represent an opportunity for the chemical industry to achieve optimal solutions faster. The results of this study illustrated the flexibility and robustness of the SA-GA algorithm to be adjusted for multi-objective systems. The limitation of this study is that the algorithms were tested on specific chemical plant systems with the intent of chemical plant optimisation. Therefore, the generalisability of the algorithms would remain in the realm of chemical plant optimisation until further case studies have been tested.

\section{Conclusion}\label{sec:conclusion}

The optimisation of the motivating examples, CPS-1 and CPS-2, has achieved result sets demonstrating the Random Forest SA-GA algorithm's robustness towards more complex systems. The significant computational efficiency gains of 1.7 times speedup for CPS-1 and 1.84 times speedup for CPS-2 and the accuracy of 90\% and 69\% of the surrogate models for the CPS-1 and CPS-2 examples respectively illustrated the robustness of the Random Forest SA-GA algorithm. The PSA system optimisation yielded results supporting that the Random Forest SA-NSGA algorithm is robust towards more complex systems, including real-world multi-objective chemical plant systems. The results illustrated the achievement of replicating the direct implementation and further achieving a 2.7 times speedup with the surrogate assisted technique with an accuracy of 91.8\% and 99.1\% for the \textit{purity} and \textit{recovery} models, respectively. These results illustrated the computational improvements obtained using a GA combined with a Machine Learning Surrogate model as a substitute for long-running simulation models. It is worth highlighting the Random Forest SA-GA and SA-NSGA algorithm's flexibility and robustness in adapting to more complex multi-objective systems. The surrogate assisted optimisation approach used in this study has proven successful across complex chemical systems and should be further verified and tested within Industry. This approach can enable faster decision making for the optimisation of chemical plant design and operations.

\section*{Conflict of interest}
The authors declare that they have no conflict of interest.
\bibliographystyle{spmpsci}      
\bibliography{ref_new}   

\end{document}